 \documentclass[final,3p,times]{elsarticle}

\usepackage{graphicx}
\usepackage{amssymb}
\usepackage[dvipsnames]{xcolor}
\usepackage{floatrow}
\usepackage{multirow}
\usepackage{booktabs}
\usepackage{longtable}
\usepackage{lscape}
\usepackage{pifont}
\usepackage[hyphens]{url}
\usepackage{hyperref}
\hypersetup{colorlinks=true,breaklinks=true}

\journal{Information and Software Technology}

\usepackage{etoolbox}
\makeatletter
\patchcmd{\ps@pprintTitle}
  {Preprint submitted to}
  {Accepted 6 March 2020 by}
  {}{}
\makeatother

\begin{document}

\begin{frontmatter}


\title{Testing and verification of neural-network-based safety-critical control software: A systematic literature review}



\author[mymainaddress,mysecondaddress]{Jin Zhang}
\ead{jin.zhang@ntnu.no}
\author[mymainaddress]{Jingyue Li\corref{mycorrespondingauthor}}
\cortext[mycorrespondingauthor]{Corresponding author}
\ead{jingyue.li@ntnu.no}
\address[mymainaddress]{Computer Science Department, Norwegian University of Science and Technology, Trondheim, Norway}
\address[mysecondaddress]{School of Information Science and Technology, Southwest Jiaotong University, Chengdu, China}

\begin{abstract}
{\bf Context:} Neural Network (NN) algorithms have been successfully adopted in a number of Safety-Critical Cyber-Physical Systems (SCCPSs). Testing and Verification (T\&V) of NN-based control software in safety-critical domains are gaining interest and attention from both software engineering and safety engineering researchers and practitioners.\\
{\bf Objective:} With the increase in studies on the T\&V of NN-based control software in safety-critical domains, it is important to systematically review the state-of-the-art T\&V methodologies, to classify approaches and tools that are invented, and to identify challenges and gaps for future studies.\\
{\bf Method:} By searching the six most relevant digital libraries, we retrieved 950 papers on the T\&V of NN-based Safety-Critical Control Software (SCCS). Then we filtered the papers based on the predefined inclusion and exclusion criteria and applied snowballing to identify new relevant papers. \\
{\bf Results:} To reach our result, we selected 83 primary papers published between 2011 and 2018, applied the thematic analysis approach for analyzing the data extracted from the selected papers, presented the classification of approaches, and identified challenges.\\
{\bf Conclusion:} 
The approaches were categorized into five high-order themes, namely, assuring robustness of NNs, improving the failure resilience of NNs, measuring and ensuring test completeness, assuring safety properties of NN-based control software, and improving the interpretability of NNs. From the industry perspective, improving the interpretability of NNs is a crucial need in safety-critical applications. We also investigated nine safety integrity properties within four major safety lifecycle phases to investigate the achievement level of T\&V goals in IEC 61508-3. Results show that correctness, completeness, freedom from intrinsic faults, and fault tolerance have drawn most attention from the research community. However, little effort has been invested in achieving repeatability, and no reviewed study focused on precisely defined testing configuration or defense against common cause failure.\\
\end{abstract}
\begin{keyword}
Software testing and verification\sep Neural network \sep Safety-critical control software \sep Systematic literature review


\end{keyword}

\end{frontmatter}


\section{Introduction}
\label{S:1}
Cyber-Physical Systems (CPSs) are systems involving networks of embedded systems and strong human-machine interactions \cite{rajkumar2010cyber}. Safety-critical CPSs (SCCPSs) are a type of CPSs that highlights the severe non-functional constraints (e.g., safety and dependability). The failure of SCCPSs could result in loss of life or significant damage (e.g., property and environmental damage). Typical applications of SCCPSs are in nuclear systems, aircraft flight control systems, automotive systems, smart grids, and healthcare systems.

In the last few years, advances in Neural Networks (NNs) have boosted the development and deployment of SCCPSs. The NN is considered the most viable approach to meet the complexity requirements of Safety-Critical Control Softwares (SCCSs) \cite{bose2007neural,ongsulee2017artificial}. In this study, we refer to NN-based SCCS as SCCS that heavily use NNs (e.g., to implement controller). For example, in the  transportation industry, deep-learning-based NNs have been widely used to developing self-driving cars \cite{bojarski2016end} and collision avoidance systems \cite{julian2016policy}. It is also worth noting that several safety incidents caused by autonomous vehicles have been presented in media, e.g., Uber car's fatal incident \cite{levin_wong_2018}, Tesla's fatal Autopilot crash  \cite{yadron_tynan_2016}, and Google's self-driving car crash \cite{lee_2016}. In addition to the safety incidents caused by failures of the autonomous system, security breaches of autonomous vehicles can potentially lead to safety issues, e.g., a demo showed that autonomous vehicles can be remotely controlled and hijacked \cite{valasek_chris_miller_charlie_2015}. How can we ensure that an SCCS containing NN technology will behave correctly and consistently when system failures or malicious attacks occur?
Increasing interest in the migration of Industrial Control Systems (ICSs) towards SCCPSs has encouraged research in the area of safety analysis of SCCPSs. Kriaa et al. \cite{kriaa2015survey} surveyed existing approaches for an integrated safety and security analysis of ICSs. The approaches cover both the design stage and the operational stage of the system lifecycle. Some approaches (such as \cite{AVEN2007745,stoneburner2006toward}) are aimed at combining safety and security techniques into a single methodology. Others (such as \cite{4638412,bieber2012security}) are trying to align safety and security techniques. These approaches are either generic, which consider both safety and security at a very high level, or model-based, which build upon the formal or semi-formal representation of the system's functions.
There are many studies that focus on the T\&V of NNs in the past decade. Several review articles \cite{taylor2003verification,hains2018towards,RN8559,RN8895} on this topic have been published. Studies \cite{taylor2003verification,van2017challenges} have reviewed methods focusing on verification and validation of NNs for aerospace systems. Studies \cite{RN8559,RN8895} are limited in automotive applications. None of these review articles have applied the Systematic Literature Review (SLR) \cite{Kitchenham07guidelinesfor} approach. 
Recently there has been more concern about Artificial Intelligence (AI) safety. The state-of-the-art advancements in the T\&V of NN-based SCCS are increasingly important; hence, there is a need to have a thorough understanding of present studies to incentivize further discussion. This study aimed to summarize the current research on \textbf{T\&V methods for NN-based control software in SCCPSs}. We have systematically identified and reviewed 83 papers focusing on the T\&V of NN-based SCCSs and synthesized the data extracted from those papers to answer three research questions. 
\begin{itemize}
\item RQ1 What are the profiles of the studies focusing on testing and verifying NN-based SCCSs?
\item RQ2 What approaches and associated tools have been proposed to test and verify NN-based SCCSs?
\item RQ3 What are the limitations of current studies with respect to testing and verifying NN-based SCCSs?
\end{itemize}

To our best knowledge, our study is the first SLR on testing and verifying  NN-based control software in SCCPSs. The results of these research questions can help researchers identify the research gaps in this area, and help industrial practitioners choose proper verification and certification methods.

The main contributions of this work are:
\begin{itemize}
\item We made a classification of T\&V approaches in both academia and industry for NN-based SCCSs.
\item We identified and proposed challenges for advancing state-of-the-art T\&V for NN-based SCCSs.
\end{itemize}

The remainder of this paper is organized as follows: In section \ref{S:2}, we define terminologies related to NN-based SCCPSs and summarize related work from academia and industry. Section \ref{S:3} describes the SLR process and the review protocol. The results of the research questions are reported in Section \ref{S:4}. Section \ref{S:5} discusses the industry practice of T\&V of NN-based SCCSs, and the threats to validity of our study. Section \ref{S:6} concludes the study.
\section{Background}
\label{S:2}
In this section, we first introduce terminology related to CPSs and modern NNs and show how NN algorithms have been used in SCCPSs. Then, we present the current state of practice of T\&V of SCCSs.
\subsection{Cyber-physical systems}
As defined in \cite{rajkumar2010cyber}, ``\textit{cyber-physical systems (CPSs) are physical and engineered systems whose operations are monitored, coordinated, controlled and integrated by a computing and communication core.}'' Several other systems, such as Internet of Things (IoTs) and ICSs have very similar features compared to CPSs, since they are all systems used to monitor and control the physical world with embedded sensor and actuator networks. In general, CPSs are perceived as the new generation of embedded control systems, which can involve IoTs and ICSs \cite{lee2015past,RN494}. 

\begin{figure}[ht]
\centering\includegraphics[width=1\linewidth]{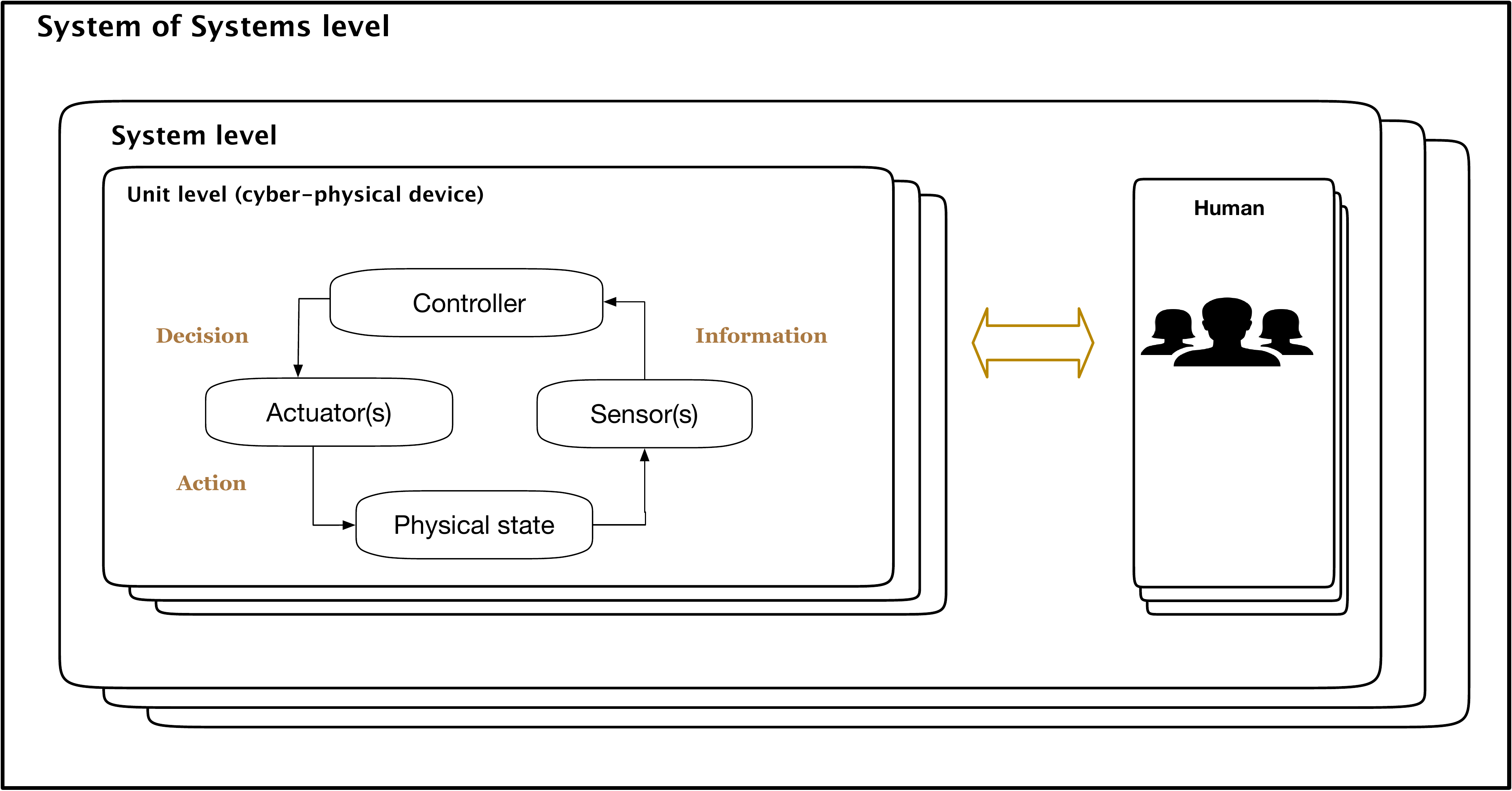}
\caption{CPS conceptual model}
\label{fig:CPS}
\end{figure}

In this SLR, we adopted the CPS conceptual model in \cite{griffor2017framework} as a high-level abstraction of CPSs to describe the different perspectives of CPSs and the potential interactions of devices and systems in a system of systems (SoS) as shown in Fig. \ref{fig:CPS}. From the perspective of unit level, a CPS at least includes one or several controllers, many actuators, and sensors. A CPS can also be a system consisting of one or more cyber-physical devices. From the SoS perspective, a CPS is composed of multiple systems that include multiple devices. In general, a CPS must contain the decision flow (from controller to actuators), information flow (from sensors to controller), and action flow (actuators impacting the physical state of the physical world).

In the context of SCCPS, safety and performance are dependent on the system (to be more specific, the controller of the system) making the right decision according to the measurement of the sensors, and operating the actuators to take the right action at the right time. Thus, verification of the process of decision-making is vital for a SCCPS.

\subsection{Modern neural networks}
The concept of “\textit{neural network}” was first proposed in 1943 by Warren McCullough and Walter Pitts \cite{mcculloch1943logical}, and Frank Rosenblatt in 1957 designed the first trainable neural network called “\textit{the Perceptron}"\cite{rosenblatt1958perceptron}. A perceptron is a simple binary classification algorithm with only one layer and output decision of “0” or “1.” By the 1980s, neural nets with more than one layer were proposed to solve more complex problems, i.e., multilayer perceptron (MLP). In this SLR, we regard multilayer NNs that emerged after the 1980s as modern NNs.

\begin{figure}[ht]
\centering\includegraphics[width=0.6\linewidth]{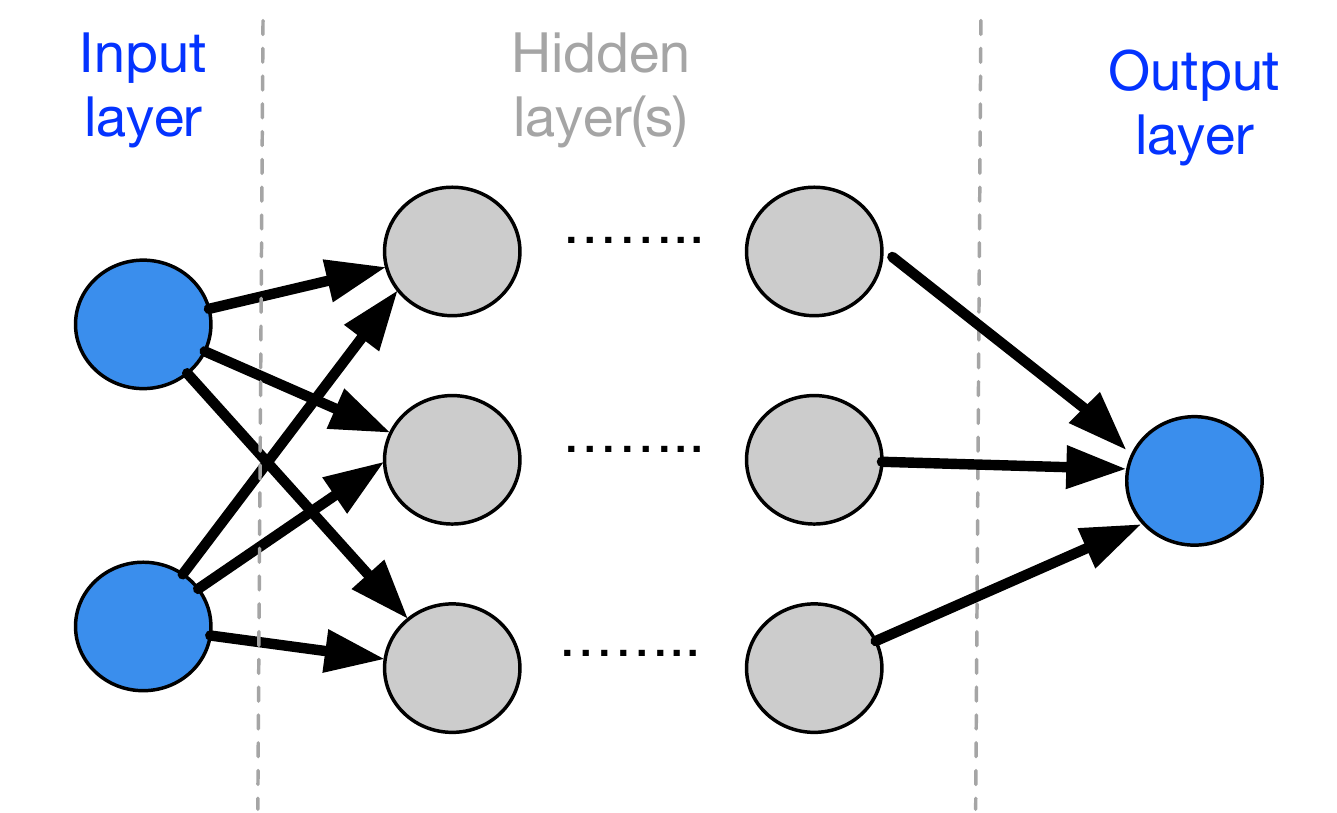}
\caption{A simplified feed-forward ANN architecture}
\label{fig:ANN}
\end{figure}

\textbf{Artificial Neural Network} \textbf{(ANN) }is the general name of computing systems designed to mimic how the human brain processes information \cite{katz2017reluplex}. An ANN is composed of a collection of interconnected computation nodes (namely “artificial neurons”), which are organized in layers. Depending on the directions of the signal flow, an ANN can have feed-forward or feedback architectures. Fig. \ref{fig:ANN} shows a simplified  feed-forward ANN architecture with multiple hidden layers. Each artificial neuron has weighted inputs, an activation function, and one output. The weights of the interconnections are adjusted based on the learning rules. There are three main models of learning rules, which are unsupervised learning, supervised learning, and reinforcement learning. The choice of learning rules corresponds to the particular learning task. The common activation functions contain sigmoid, hyperbolic tangent, radial bases function (RBF), and piece-wise linear transfer function, such as Rectified Linear Unit (ReLU) \cite{kruse2013multi}. In a word, an ANN can be defined by three factors: the interconnection structure between different layers, activation function type, and procedure for updating the weights. 

\textbf{Multi-Layer Perceptron (MLP \cite{lecun2015deep})} represents a class of feed-forward ANN. An MLP consists of an input layer, one or several hidden layers, and an output layer. Each neuron of MLP in one layer is fully connected with every node in the following layer. An MLP employs a back-propagation technique (which belongs to supervised learning) for training.

\textbf{Convolutional Neural Network (CNN \cite{lecun1998gradient}) }is a special type of multi-layer NN with one or more convolutional layers. A convolutional layer includes ``\textit{several feature maps with different weight vectors. A sequential implementation of a feature map would scan the input image with a single unit that has a local receptive field, and store the states of this unit at corresponding locations in the feature map. This operation is equivalent to a convolution, followed by an additive bias and squashing function, hence the name convolutional network}''\cite{lecun1998gradient}. CNNs are superior for processing two-dimensional data (particular camera images) because of the convolution operations, which are capable of detecting features in images. CNNs are now widely applied to develop partially-autonomous and fully-autonomous vehicles.

\textbf{Deep Neural Networks (DNNs \cite{van2018artificial})} represent an ANN with multiple hidden layers between the input and output layers. DNNs (e.g., a MLP with more than three layers or a CNN) differ from shallow NNs (e.g., a three-layer MLP) in the number of layers, the activation functions that can be employed, and the arrangement of the hidden layer. Compared to shallow NNs, DNNs can be trained more in-depth to find patterns with high performance even for complex nonlinear relationships. 

An NN could be trained offline or online. An NN trained offline means it only learns during development. After training, the weights of the NN will be fixed and the NN will act deterministically. Therefore static verification methods could be possible. In contrast, online training will allow the NN to keep learning and evolving during operation, which requires run-time verification methods. In some applications, such as the Intelligent Flight Control System developed by NASA \cite{taylor2003verification}, both offline and online training strategies are employed to meet the system requirements.

NNs are fundamentally different with algorithmic programs, but a formal development methodology can still be derived for an NN system. Development process of an NN system can include six phases \cite{836192}:

\begin{enumerate}
\item Formulation of requirements and goals;
\item Selection of training and test data sets;
\item Selection of the NN architecture;
\item Training of the network;
\item Testing of the network; and
\item Acceptance and use by the customer.
\end{enumerate}

Like \cite{836192}, Falcini et al. introduced  a similar development lifecycle for DNNs in automotive software \cite{7927925} and proposed a W-model integrated data development with standard software development to highlight the importance of data-driven in DNN development. Falcini et al. \cite{7927925} also summarized that the DNN's functional behavior depends on both its architecture and its learning outcome through training.

\subsection{The trends of using NN algorithm in SCCPSs}
From 1940s automated range finders (developed by Norbert Wiener for anti-aircraft guns) \cite{heilbron2003oxford} to today's self-driving cars, AI, especially NN algorithms, is widely applied in both civilian (e.g., autonomous cars) and military domains (e.g., military drones). Boosted by the advances of AI, state-of-the-art CPSs can plan and execute  more and more complex operations with less human interaction. Here we present the applications of NNs in the following four representative SCCPSs.

\subsubsection{Autonomous cars}
For automobile, the Society of Automotive Engineers (SAE) proposed six levels of autonomous driving \cite{RN547}. A level 0 vehicle has no autonomous capabilities, and the human driver is responsible for all aspects of the driving task. For level 5 vehicle, the driving tasks are only managed by the autonomous driving system. When developing autonomous vehicles targeting a high level of autonomy, one industry trend is to use DNNs to implement vehicle control algorithms. The deep-learning-based approach enables vehicles to learn meaningful road features from raw input data automatically and then output driving actions. The so-called end-to-end learning approach can be applied to resolve complex real-world driving tasks. When using deep-learning-based approaches, the first step is to use a large number of training data sets (images or other sensor data) to train a DNN. Then a simulator is used to evaluate the performance of the trained network. After that, the DNN-based autonomous vehicle will be able to \textit{``execute recognition, prediction, and planning"} driving tasks in diverse conditions \cite{kriaa2015survey}. Nowadays, CNNs are the most widely adopted deep-learning model for fully autonomous vehicles \cite{julian2016policy,levin_wong_2018,yadron_tynan_2016,lee_2016}. NVIDIA introduced an AI supercomputer for autonomy \cite{RN577}. The development flow using NVIDIA DRIVE PX includes four stages: 1) data acquisition to train the DNN, 2) deployment of the output of a DNN in a car, 3) autonomous application development, and 4) testing in-vehicle or with simulation.

One essential characteristic of deep-learning-based autonomy is that the decision-making part of the vehicle is almost a black box. This means that in most cases, we as human drivers must trust the decisions made by the deep-learning algorithms without knowing exactly why and how the decisions are made.
\subsubsection{Industrial control systems}
Industrial Control System (ICS) is the general term for control systems, also called Supervisory Control and Data Acquisition (SCADA) systems. ICSs make decisions based on the specific control law (such as lookup table and non-linear mathematical model) formulated by human designers. In contrast to the classical design procedure of control law, reinforcement-learning-based approaches learn the control law simply from the interaction between the controller and the process, and then incrementally improving control behavior. Such approaches and NNs have been used in process control two decades ago \cite{RN8998}. Concerning the recent progress in AI and the success of DNNs in making complex decisions, there are high expectations for the application of DNNs in ICSs. For instance, DNNs and reinforcement learning can be combined to develop continuous control \cite{RN8999}. Spielberg et al. extended the work in \cite{RN8999} to design control policy for  process control \cite{RN8997}. Even though the proposed approach in \cite{RN8997} is only tested on linear systems, it shows a practical solution for applying DNNs in non-linear ICSs. 
\subsubsection{Smart grid systems}
 The smart grid is designed as the next generation of electric power system, dependent on information communications technology (ICT). There is tremendous initiative of research activities in automated smart grid applications, such as FLISR (which is a smart grid multi-agent automation architecture based on decentralized intelligent decision-making nodes) \cite{RN8991}. NNs have been considered for solving many pattern recognition and optimization problems, such as fault diagnosis \cite{RN8989}, and control and estimation of flux, speed \cite{bose2007neural}, and economical electricity distribution to consumers. MLP is one of the most commonly used topology in power electronics and motor drives \cite{bose2007neural}. 
\subsubsection{Healthcare}
Medical devices is another emerging area where research and industry practitioners are seeking to integrate AI technologies to improve accuracy and automation. ANNs and other machine learning approaches have been proposed to improve the control algorithms for diabetes treatment in recent decades \cite{RN8993,RN8994}. In 2017, an AI-powered device for automated and continuous delivery of basal insulin (named MiniMed 670G system \cite{RN8995}) was approved by the U.S. Food and Drug  Administration. In the same year, it was reported that GE Healthcare had integrated the NVIDIA AI platform into their computerized tomography scanner to improve speed and accuracy for the detection of liver and kidney lesions \cite{RN8992}. Using deep learning solutions, such as CNNs, in the medical computing field has proven to be effective since CNNs have excellent performance in object recognition and localization in medical images \cite{RN8996}.
\subsection{Testing and verification of safety-critical control software}
 IEC 61508 and ISO 26262 are two standards highly relevant to the T\&V of SCCS. IEC 61508 is an international standard concerning \textit{Functional safety of electrical/electronic/programmable electronic safety-related systems}. It defines four safety integrity levels (SILs) for safety-critical systems \cite{RN9010}. The higher the SIL level a SCCPS requires, the more time and effort for verification are needed. In IEC 61508, formal methods are highly recommended techniques for verifying high SIL systems. Because formal methods can be used to construct the specification and provide a mathematical proof that the system matches some formal requirements, this is quite a strong commitment for the correctness of a system. 

ISO 26262, titled \textit{Road vehicles – functional safety}, is an international standard for the functional safety of electrical and/or electronic systems in production automobiles \cite{ISO26262}. Besides using classical safety analysis methods such as Fault Tree Analysis (FTA) and Failure Mode and Effects Analysis (FMEA), ISO 26262 explicitly  states that  the production of a safety case is mandated to assure system safety. It defines a safety case as “an argument that the safety requirements for an item are complete and satisfied by evidence compiled from work products of the safety activities during development” \cite{ISO26262}. 

The development of suitable approaches, which can verify the system behavior and misbehavior of a SCCPS is always challenging. Not to mention that the architecture of NNs (especially DNNs) makes it even harder to decipher how the algorithmic decisions were made. The current version of IEC 61508 is not applicable for the verification of NN-based SCCSs because AI technologies are not recommended there. The latest version of ISO 26262 and its extension, ISO/PAS 21448, which is also known as safety of the intended functionality (SOTIF) \cite{RN8974}, will likely provide a way to handle the development of autonomous vehicles. However, SOTIF will only provide guidelines associated with SAE Level 0--2 autonomous vehicles \cite{RN8978}, which are not ready for the verification of NN-based autonomous vehicles.

In practice, in order to reduce test and validation costs, high-fidelity simulation is a commonly used approach in the automotive domain. The purpose of using a simulator is to predict the behavior of an autonomous car in a mimicked environment. NVIDIA and Apollo distributed their high-fidelity simulation platforms for testing autonomous vehicles. CARLA \cite{RN9011} and Udacity's Self-Driving Car Simulator \cite{Udacity} are two popular open-source simulators for autonomous driving research and testing.
\section{Research method}
\label{S:3}
We conducted our SLR by following the SLR guidelines in \cite{Kitchenham07guidelinesfor} as well as consulting other relevant guidelines in \cite{RN8908} and \cite{RN8926,RN510}. Our review protocol consisted of four parts: 1) search strategy, 2) inclusion and exclusion criteria, 3) selection process, and 4) data extraction and synthesis.
\subsection{Search strategy}
Based on guidelines provided in \cite{Kitchenham07guidelinesfor}, we use the Population, Intervention, Outcome, Context (PIOC) criteria to formulate search terms. In this SLR, 
\begin{itemize}
\item The population should be an application area (e.g., general CPS) or specific CPS (e.g., self-driving car). 
\item The intervention is methodology, tools and technology that address system/component testing or verification. 
\item The outcome is the improved safety or functional safety of CPSs. 
\item The context is the NN-based SCCPSs in which the T\&V take place. 
\end{itemize}

\begin{figure}[ht]
\centering\includegraphics[width=1\linewidth]{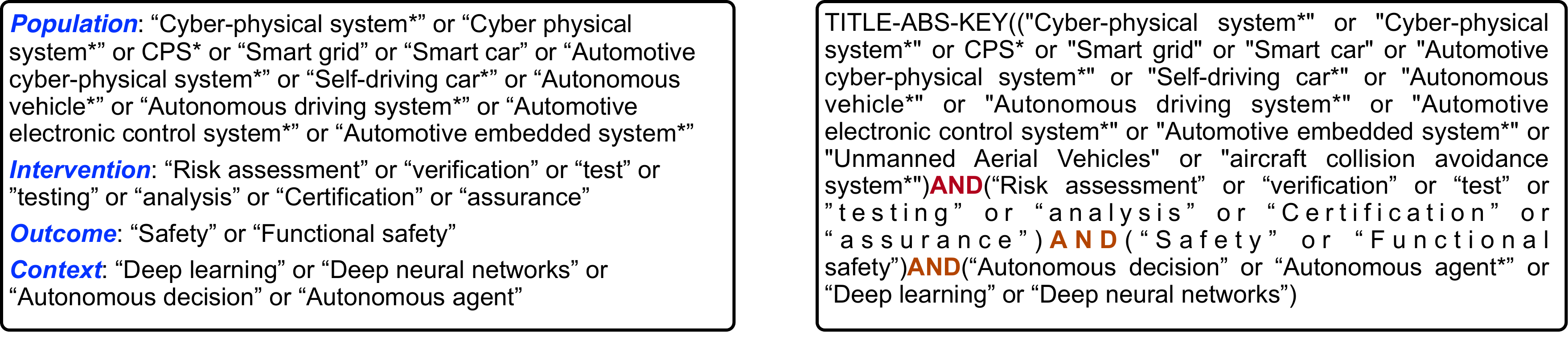}
\caption{Search terms}
\label{fig:serachterms}
\end{figure}

Fig. \ref{fig:serachterms} shows the search terms formulated based on the PIOC criteria. We first used these search terms to run a series of trial searches and verify the relevance of the resulting papers. We then revised the search string to form the final search terms. The final search terms were composed of synonyms and related terms. 

We executed automated searches in six digital libraries, namely, Scopus, IEEE Xplore, Compendex EI, ACM Digital library, SpringerLink, and Web of Science (ISI). 
\subsection{Inclusion and exclusion criteria}
Table \ref{table:criteria} presents our inclusion and exclusion criteria. We set three inclusion criteria to restrict the application domain, context, and outcome type. We excluded papers that were not peer-reviewed, such as keynotes, books, and dissertations, and papers not written in English. It should be clarified that, unlike most other SLR studies, we did not directly exclude short papers (less than six pages), work-in-progress papers, and pre-print papers. The reason is that this research area is far from mature, so many initial thoughts or in-progress papers are still valuable to review.
\begin{table}[ht]
\centering
\resizebox{\textwidth}{!}{
\begin{tabular}{*{2}{l}}
\hline
\multicolumn{2}{l}{\textbf{Inclusion criteria}}\\
\hline
I1 & The paper must have a context in SCCPSs, either in general or in a specific application domain\\
I2 & The paper must be aimed at testing/verification approaches for NN-based SCCSs\\
I3 & The paper must be aimed at modern neural networks\\
\hline\hline
\multicolumn{2}{l}{\textbf{Exclusion criteria}}\\
\hline
E1&Papers not peer-reviewed  \\
E2&Not written in English \\
E3&Full-text is not available  \\
E4 &Not relevant to modern neural networks  \\
\hline
\end{tabular}}
\caption{Inclusion and Exclusion criteria}
\label{table:criteria}
\end{table}
\subsection{Selection process}
\begin{figure}[ht]
\centering\includegraphics[width=0.8\linewidth]{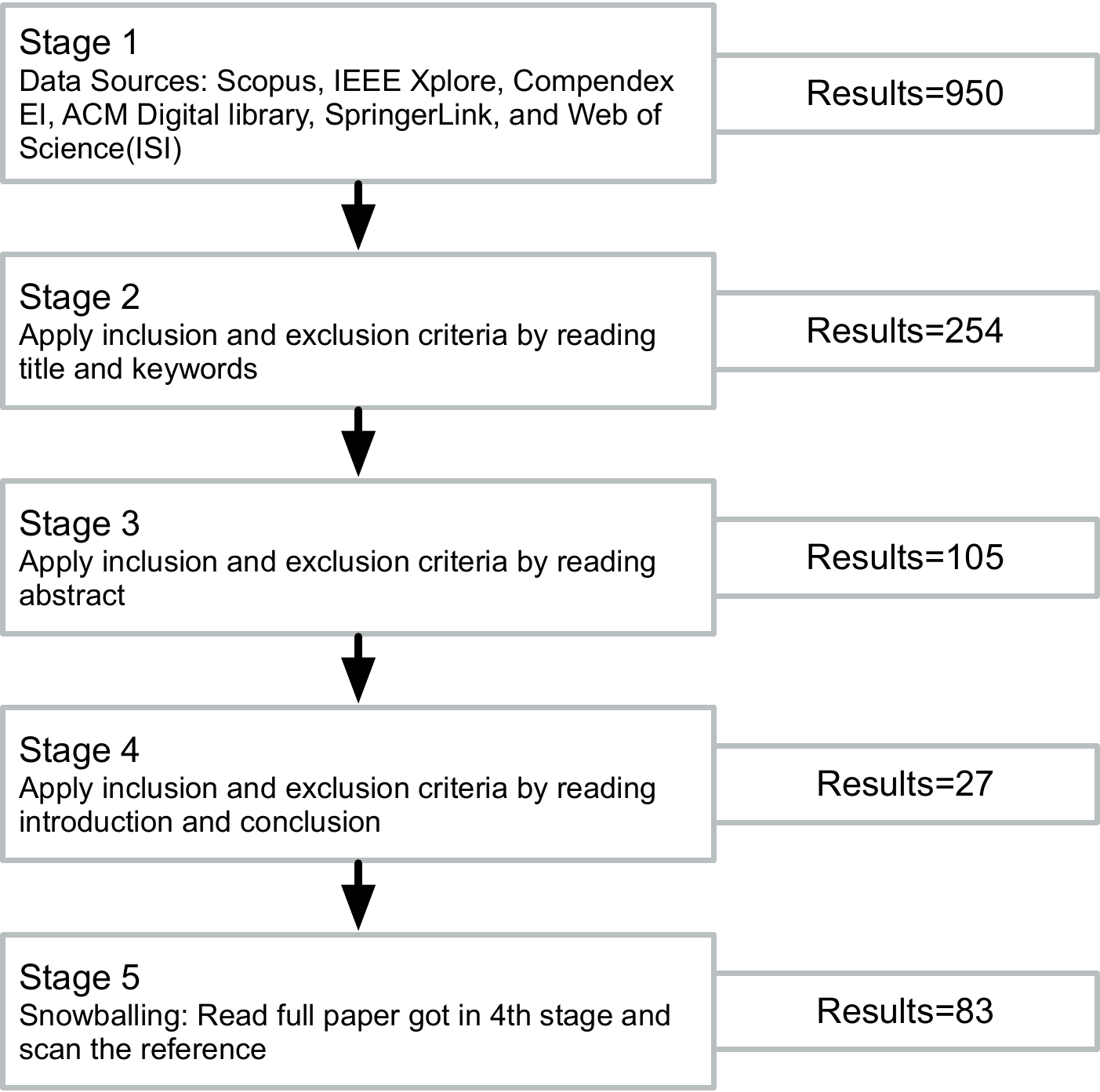}
\caption{Search process}
\label{fig:serachprocess}
\end{figure}

We used the inclusion and exclusion criteria to filter the papers in the following steps. We covered papers from January 2011 to November 2018. Fig. \ref{fig:serachprocess} shows the whole search and filtering process.

\textbf{Stage 1:} Ran the search string on the six digital libraries and retrieved 1046 papers. After removing those duplicated papers, we had 950 papers. 

\textbf{Stage 2:} Excluded studies by reading title and keywords. If it was not excluded simply by reading titles and keywords, the paper was kept for further investigation. At the end of this stage, we selected 254 papers. 

\textbf{Stage 3:} Further filtered the papers by reading abstracts and found 105 potential papers with high relevance to the research goal of our SLR.

\textbf{Stage 4:} Read the introduction and conclusion to decide on selection. We recorded the reasons for exclusion for each excluded paper. We excluded the papers that were irrelevant, or whose full texts were not available. Furthermore, we critically examined the quality of primary studies to exclude those that lacked sufficient information. We ended up with 27 papers.

\textbf{Stage 5:} Read full text of the selected studies from the fourth stage, applied snowballing by scanning the reference of the selected papers. The snowballing process can be implemented in two directions: backwards (which means scanning the references of a selected paper and find any other relevant papers published before the selected paper), and forwards (which means checking if any other relevant paper was published after the selected paper and cited the selected paper). In our SLR, we adapted mainly backward snowballing to include additional papers. To limit the scope of the snowballing, we covered only references published between 2011 and 2018. From snowballing, we  found 56 new relevant papers. 

Finally, we selected 83 papers as primary studies for detailed analysis. We listed all of the selected studies in Appendix A. The first author conducted the selection process with face-to-face discussions with the second author. The second author performed a cross-check of each step and read all the final selected papers to confirm the selection of the papers. 

\subsection{Data extraction and synthesis}
\textbf{Data Extraction:} We extracted two kinds of information from the selected papers. To answer RQ1, we extracted information for statistical analysis, e.g., publication year and research type. To answer RQ2 and RQ3, we collected information to identify key features (such as research goal, technique and tools, major contribution and limitation) of T\&V approaches. 

\textbf{Synthesis:} We used descriptive statistics to analyze the data for answering RQ1. To answer RQ2 and RQ3, we analyzed the data using the qualitative analysis method by following the five steps of thematic analysis \cite{RN8963}: 1) extracting data, 2) coding data, 3) translating codes into themes, 4) creating a model of higher-order themes, and 5) assessing the trustworthiness of the synthesis.

\section{Result}
\label{S:4}
\subsection{RQ1. What are the profiles of the studies focusing on testing and verifying NN-based SCCSs?}
\label{SS:1}
\begin{figure}[ht]
\centering\includegraphics[width=0.8\linewidth]{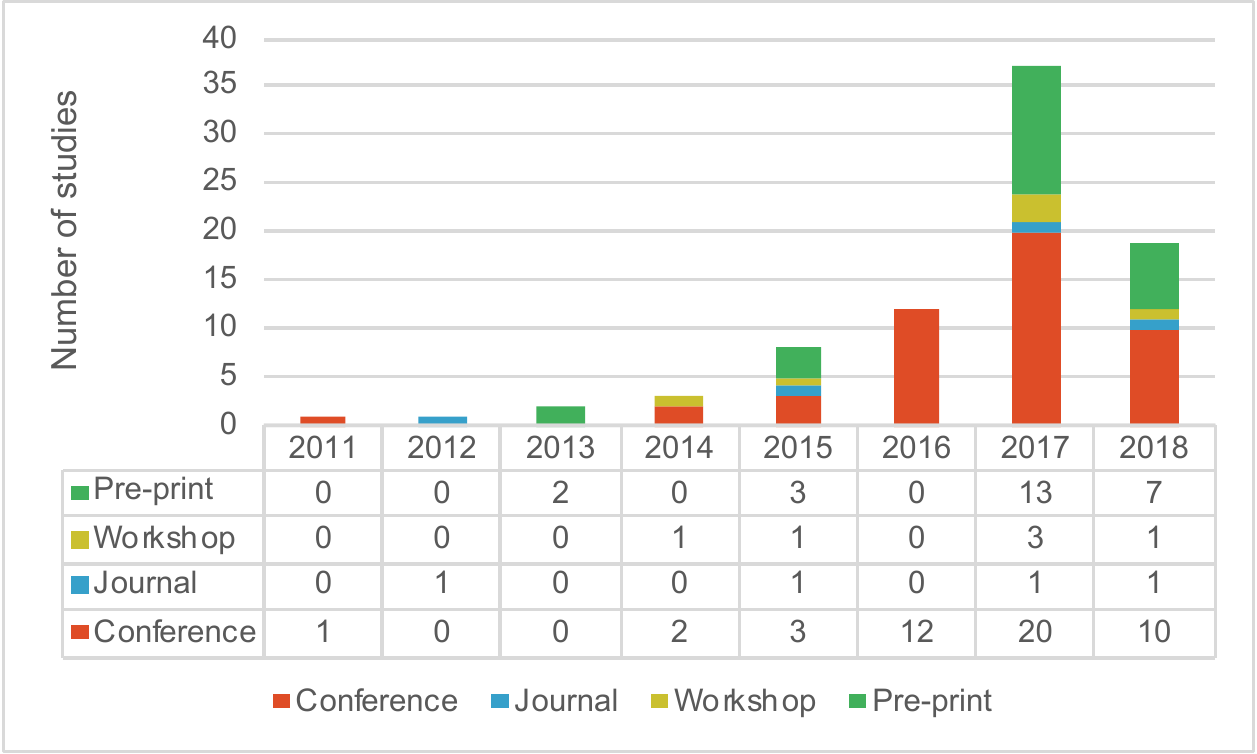}
\caption{Publication year and types of selected papers}
\label{fig:yeartype}
\end{figure}

\textbf{Studies distributions:} Fig. \ref{fig:yeartype} shows the distribution of selected papers based on publication year and the types of work. There has been 68 papers (81.9\%) published since 2016, indicating that researchers are paying more attention to the T\&V of NN-based SCCSs. Conference was the most popular publication type with 48 papers (57.8\%), followed by pre-print (25 papers, 30.1\%), workshop (6 papers, 7.2\%), and journal (4 papers, 4.8\%). 

We also investigated the geographic distribution of the reviewed studies. It allowed us to identify which countries are leading the research in this domain. We considered a study to be conducted in one country if the affiliation of at least one author is in that country. Moreover, the involvement of industry would be an indicator of industry's interest in this domain. We classified the reviewed papers as industry if at least one author came from industry or the study used real-world industrial systems to test/verify the proposed approach. A paper would be categorized as academia if all authors came from academia. It shows that researchers based in the USA have been involved in the most primary studies for testing or verification of NN-based SCCSs with 56 publications, followed by the researchers based in Germany and the UK with 10 and 9 publications, respectively. It is worth noting that 47 of 83 (56.6\%) publications have involvement from industry.

\textbf{Research types:} We classified the selected papers based on the criteria proposed by Kai et al. \cite{RN8908} (See Table \ref{tab:criterion}). According to Table \ref{tab:criterion}, the research type of the paper is governed by rules (i.e., R1-R6). Each rule is a combination of several conditions. The six research types (i.e., evaluation research, solution proposal, validation research, philosophical papers, opinion papers, and experience papers) correspond to R1-R6, respectively. For example, both evaluation research (corresponding to R1) and validation research (corresponding to R4) must present empirical evaluation. The difference between evaluation and validation research is that validation is not used in practice (e.g., experimental or simulation-based approaches), whereas evaluation studies should be conducted in a real-world context. Solution proposal means that it has to propose a new solution that may or may not be used in practice. We found that evaluation and validation research are the majority of the selected papers, corresponding to 31.3\% (26 papers) and 61.4\% (51 papers) of the selected papers, respectively. The low percentage of the  solution proposal (6 papers) was not surprising because a majority of the reviewed papers presented and demonstrated their T\&V approaches through academic and industrial case studies, simulation, and controlled experiments. The other three types of research papers (i.e., philosophical papers, opinion papers, and experience papers) do not exist in selected studies because we only included papers that aimed at testing/verification approaches (refer to inclusion criteria I2).

\begin{table}[ht]
\centering
\caption{Research type classification (T = True, F = False, \ding{108} = irrelevant or not applicable, R1–R6 refer to rules).}
\label{tab:criterion}
\resizebox{0.8\textwidth}{!}{%
\begin{tabular}{lllllll}
\hline
 & R1 & R2 & R3 & R4 & R5 & R6 \\
 \hline
\textit{Conditions} &  &  &  &  &  &  \\
Used in practice & T & \ding{108} & T & F & F & F \\
Novel solution & \ding{108} & T & F & \ding{108} & F & F \\
Empirical evaluation & T & F & F & T & F & F \\
Conceptual framework & \ding{108} & \ding{108} & \ding{108} & \ding{108} & T & F \\
Opinion about something & F & F & F & F & F & T \\
Authors' experience & \ding{108} & \ding{108} & T & \ding{108} & F & F \\
 &  &  &  &  &  &  \\
\textit{Decisions} &  &  &  &  &  &  \\
Evaluation research &\ding{51}& \ding{108} & \ding{108} & \ding{108} & \ding{108} & \ding{108} \\
Solution proposal & \ding{108} & \ding{51} & \ding{108} & \ding{108} & \ding{108} & \ding{108} \\
Validation research & \ding{108} & \ding{108} & \ding{108} & \ding{51} & \ding{108} & \ding{108} \\
Philosophical papers & \ding{108} & \ding{108} & \ding{108} & \ding{108} & \ding{51} & \ding{108} \\
Opinion papers & \ding{108} & \ding{108} & \ding{108} & \ding{108} & \ding{108} & \ding{51} \\
Experience papers & \ding{108} & \ding{108} & \ding{51} & \ding{108} & \ding{108} & \ding{108}\\
\hline
\multicolumn{1}{l}{Note: Reprinted from \cite{RN8908},Copyright 2015 by the Elsevier.}
\end{tabular}}
\end{table}

\textbf{Application domains:} We analyzed the application domain of selected studies to provide useful information for researchers and practitioners who are interested in the domain-specific aspects of the approaches. The results are shown in Table \ref{table:applicatondomain}. We found that considerable effort is now being put into using NN algorithms to accomplish control logic for general purpose (59 papers, 71.1\%), automotive CPSs, such as autonomous vehicles (13 papers, 15.7), and autonomous aerial systems, such as airborne collision avoidance systems for unmanned aircrafts (5 papers, 6\%).
\begin{table}[ht]
\centering
\resizebox{0.5\textwidth}{!}{%
\begin{tabular}{*{3}{l}}
\hline
\textbf{Application domain} & \textbf{No. of studies} \\
\hline
General SCCPSs &59\\
Automotive CPSs  & 13 \\
Autonomous aerial systems & 5 \\
Robot system & 5\\
Health care & 1  \\
\hline
\end{tabular}}
\caption{Distribution of application domains of the selected studies}
\label{table:applicatondomain}
\end{table}
\subsection{RQ2. What approaches and associated tools have been proposed to test and verify NN-based SCCSs?}
\label{SS:2}
\begin{figure}[ht]
\centering\includegraphics[width=0.8\linewidth]{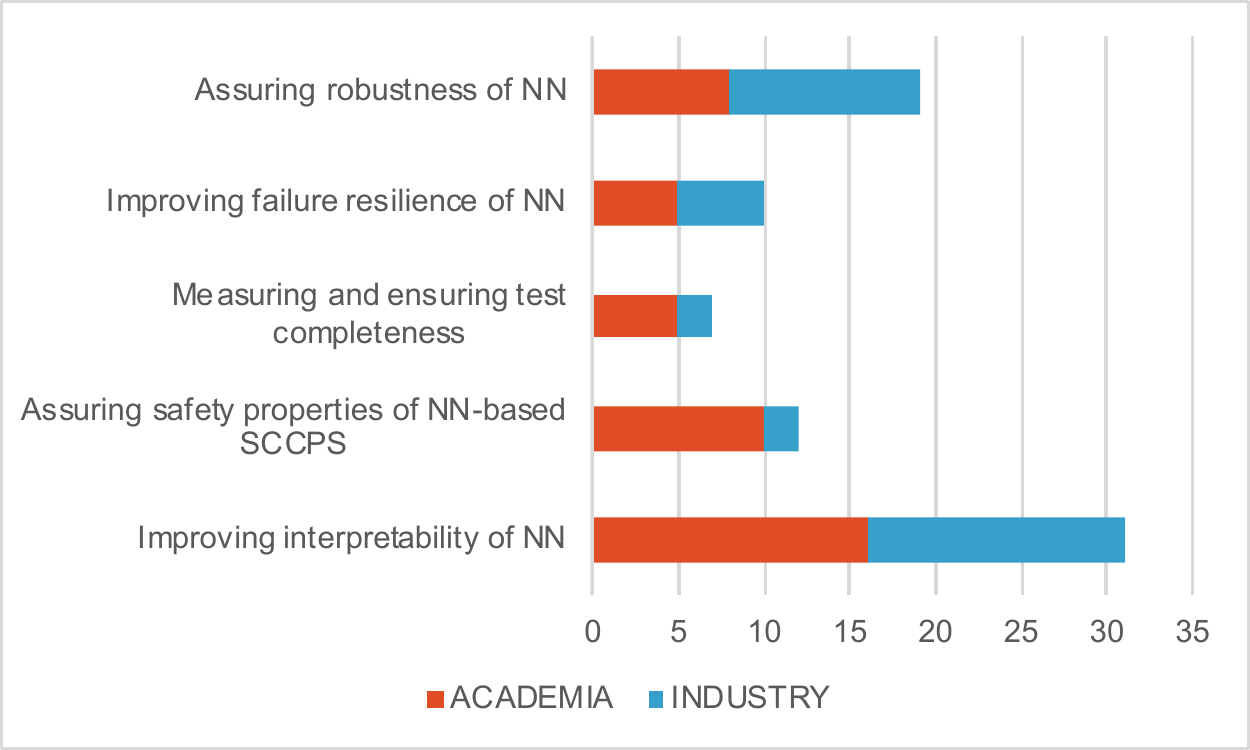}
\caption{Comparing the interests difference of academia and industry}
\label{fig:RQ1_AIfocus}
\end{figure}

As 4 of the 83 papers focused mainly on high-level ideas and concepts without presenting detailed approaches or tools, we did not include them to answer RQ2. For the remaining 79 out of 83 (95.2\%) papers, we applied the thematic analysis approach \cite{RN8963} and identified five high-order themes and some sub-themes. Some papers contain more than one themes. In order to balance the accuracy and the simplicity of categorization, we decided to assign each study only one category based on its major contribution. Table \ref{table:classification} presents the themes, sub-themes, and the corresponding papers. Fig. \ref{fig:RQ1_AIfocus} compares the interests difference of academia and industry for the five identified themes. 

\begin{table}[ht]
\centering
\resizebox{\textwidth}{!}{
\begin{tabular}{p{5cm}p{8cm}p{7cm}l}
\hline
\textbf{Themes} & \textbf{Sub-themes} & \textbf{Papers} & \textbf{\#}\\
\hline
\multirow{4}{*}{Assuring robustness of NNs} &Understanding the characteristics and impacts of adversarial examples& \cite{S7},\cite{S15},\cite{S23},\cite{S26},\cite{S32}, \cite{S37}, \cite{S40} &17\\
&Detect adversarial examples& \cite{S3},\cite{S9}, \cite{S11}, \cite{S19}, \cite{S20}, \cite{S29} &\\
&Mitigate impact of adversarial examples& \cite{S12}, \cite{S16} & \\
&Improving robustness of NNs through using adversarial examples & \cite{S21}, \cite{S42} &\\
 \hline
Improving failure resilience of NNs& \cite{S1},\cite{S2},\cite{S4},\cite{S44},\cite{S14},\cite{S18},\cite{S28},\cite{S43},\cite{S54},\cite{S55},\cite{S22}&&11\\
 \hline
Measuring and ensuring test completeness& \cite{S5},\cite{S8},\cite{S10},\cite{S49},\cite{S51},\cite{S52},\cite{S53} &&7 \\
 \hline
Assuring safety properties of NN-based CPSs& \cite{S6},\cite{S17},\cite{S27},\cite{S30},\cite{S31},\cite{S33},\cite{S34},\cite{S35},\cite{S36}, \cite{S45},\cite{S46},\cite{S50},\cite{S56}&&13\\
 \hline
\multirow{3}{*}{\shortstack{Improving interpretability \\ of NNs}}
& Understand how a specific decision is made &\multicolumn{1}{l}{\begin{tabular}[c]{@{}l@{}}\cite{S24},\cite{S58},\cite{S59},\cite{S60},\cite{S61},\cite{S64},\cite{S65},\cite{S66},\\ \cite{S68},\cite{S70},\cite{S72},\cite{S73},\cite{S74},\cite{S75},\cite{S76}, \cite{S77},\\ \cite{S78},\cite{S79},\cite{S83}\end{tabular}} &\\
& Facilitate understanding of the internal logic of NNs & \cite{S13},\cite{S62},\cite{S63},\cite{S69},\cite{S71},\cite{S80} &{31}\\
&Visualizing internal layers of NNs to help identify errors in NNs & \cite{S47},\cite{S48},\cite{S57},\cite{S67},\cite{S81},\cite{S82} &\\
 \hline
 \end{tabular}}
\caption{A classification of approaches to test and verify NN-based SCCSs}
\label{table:classification}
\end{table}

\subsubsection{CA1: Assuring robustness of NNs}
\label{SSS:3}
One high-order theme of the studies is to assure the robustness of NNs. Robustness of an NN is its ability to cope with erroneous inputs. The erroneous inputs can be an adversarial example (i.e., an input that adds small perturbation intentionally to mislead classification of an NN), or benign but wrong input data. Methods under this theme can be further classified into four sub-themes.

\textbf{Studies focusing on understanding the characteristics and impacts of adversarial examples.} Some studies tried to identify the characteristics and impacts of adversarial examples. The study \cite{S15} found the characteristics, such as the linear nature, of adversarial examples. The study \cite{S26} measured the impact of adversarial examples by counting their frequencies and severities. Nguyen et al.  \cite{S7} found that a CNN trained on ImageNet \cite{5206848} is vulnerable to adversarial examples generated through Evolutionary Algorithms (EAs) or gradient ascent. 

A few other studies, such as \cite{S23, S32, S37, S40}, tried to understand the characteristics of robust NNs. Cisse et al. \cite{S32} introduced a particular form of DNN, namely Parseval Networks, that is intrinsically robust to adversarial noise. Gu et al.  \cite{S40} concluded that some training strategies, for example, training using adversarial examples or imposing contractive penalty layer by layer, are robust to certain structures of adversarial examples (e.g., inputs corrupted by Gaussian additive noises or blurring). Higher-confidence adversarial examples (i.e., adversarial instances that are extremely easy to classify into the wrong category) were used to evaluate the robustness of the state-of-the-art NN in \cite{S37} and the robot-vision system in \cite{S23}. 

\textbf{Studies focusing on methods to detect adversarial examples.} Detecting adversarial examples that are already inserted into training or testing data set are the primary targets of \cite{S3, S11, S19, S20, S29}. Wicker et al. \cite{S3} and \cite{S20} formulated the adversarial examples detection as a two-player stochastic game and used the Monte Carlo Tree Search to identify adversarial examples. Reuben \cite{S11} applied density estimates, and Bayesian uncertainty estimates to detect adversarial samples. Xu et al. \cite{S19} proposed a feature squeezing framework to detect adversarial examples, which are generated by seven state-of-the-art methods. According to \cite{S19}, an advantage of feature squeezing is that it did not change the underlying model. Therefore, it can easily be integrated with other defenses methods. Metzen et al. \cite{S29} embedded DNNs with a subnetwork (called ``detector") to detect adversarial perturbations. The Deepsafe presented in \cite{S9} used clustering technology to find candidate-safe regions first and then verified whether the candidates were safe using counter-examples as a proof.

\textbf{Studies focusing on methods to mitigate impact of adversarial examples.} Papemot et al. \cite{S12} adopted defensive distillation as a defense strategy to train DNN-based classifiers against adversarial examples. However, several powerful attacks have been proposed to defeat defensive distillation and have demonstrated that defensive distillation does not actually eliminate adversarial examples \cite{S37}. Papemot et al. \cite{S16} revisited defensive distillation and proposed a more effective way to defend against three recently discovered attack strategies, i.e., the Fast Gradient Method (FGM) \cite{S15}, the Jacobian Saliency Map Approach (JSMA) \cite{RN8964}, and the AdaDelta optimization strategy (AdaDelta) \cite{S37}.

\textbf{Studies focusing on increasing robustness of NNs through using adversarial examples.} In studies \cite{S21} and \cite{S42}, the authors proposed methods to leverage adversarial training (e.g., generating a large amount of adversarial examples and then training the NN not to be fooled by these adversarial examples) to increase the robustness of NNs. 
\subsubsection{CA2: Improving failure resilience of NNs }
\label{SSS:4}
Studies under this theme focused on improving the resilience of NNs, so that the NN-based CPSs are more tolerant of possible hardware and software failures. 

Studies \cite{S4, S14, S18} investigated error detection and mitigation mechanisms, while studies \cite{S44, S43} focused on understanding error propagation in DNN accelerators. Vialatte et al.\cite{S4} demonstrated that faulty computations can be addressed by increasing the size of NNs. Santos et al. \cite{S14} proposed an algorithm-based fault tolerance (ABFT) strategy to detect and correct radiation-induced errors. In \cite{S18}, a binary classification algorithm based on temporal and stereo inconsistencies was applied to identify errors caused by single frame object detectors. Li et al. \cite{S44} developed a general-purpose GPU (GPGPU) fault injection tool \cite{lu2015llfi} to investigate error propagation patterns in twelve GPGPU applications. Later, Li et al.  revealed that the error resilience of DNN accelerators depends on ``\textit{the data types, values, data reuse, and the types of layers in the design \cite{S43}}''. Based on this finding, they devised guidelines for designing resilient DNN systems and proposed two DNN protection techniques, namely Symptom-based Error Detectors (SED) and Selective Latch Hardening (SLH) to mitigate soft errors that are typically caused by high-energy particles in hardware systems \cite{RN8965}. 

Mhamdi et al. explored error propagation mechanism in an NN \cite{S28}, and they theoretically and empirically proved that the key parameters that can be used to estimate the robustness of an NN are: \textit{``Lipschitz coefficient of the activation function, distribution of large synaptic weights, and depth of the network''}. The study \cite{S54} characterized the faults propagation through an open-source autonomous vehicle control software (i.e., openpilot) to assess the failure resilience of the system. The Systems-Theoretic Process Analysis (STPA) \cite{leveson2011engineering} hazard analysis technique was used to guide fault injection. Existing work in \cite{S54} showed that STPA is suited for an in-depth identification of unsafe scenarios, and thus, the fault injection space was reduced.

Based on the diversified redundancy strategies, the study \cite{S55} developed diverse networks in the context of different training data sets, different network parameters, and different classification mechanisms to strengthen the fault tolerance of the DNN architecture.

Studies \cite{S1, S2} tried to improve computation efficiency without compromising error resilience. Studies \cite{S1, S2} also predicted the error resilience of DNN accelerators to make reconfigurable NN accelerators. The study \cite{S1} demonstrated a more accurate neuron resilience assignment than the state-of-the-art techniques and provided the possibility of moving parts of the neuron computations to unreliable hardware at the given quality constraint. Zhang et al. \cite{S2} proposed a framework to increase efficiency of computation by approximating the computation of certain less critical neurons. Daftry et al. \cite{S22} provided an interesting idea about ``how to enable a robot to know when it does not know?'' The idea of \cite{S22} is to utilize the resulting features of the controller, which are learned from a CNN to predict the failure of the controller, and then let the system self-evaluate and decide whether to execute or discard an action.
\subsubsection{CA3: Measuring and ensuring test completeness}
\label{SSS:5}
The approaches and tools under this theme aim to ensure good coverage when testing NNs. The testing approaches include black-box testing (i.e., focusing on whether the tests cover all possible usage scenarios), white-box testing (i.e., focusing on whether the tests cover every neuron in the NN), and metamorphic testing, which focuses on both test case generation and result verification \cite{Chen:2018:MTR:3177787.3143561}. 

O'Kelly et al.\cite{S5} proposed methods to ensure good usage coverage through first making a formal Scenario Description Language (SDL) to create driving scenarios, and then translating the scenarios to a specification-guided automatic test generation tool named S-TALIRO to generate and run the tests. Raj et al. \cite{S49} proved the possibility of speeding up the generation of new and interesting counterexamples by introducing fuzzing patterns obtained from an unrelated DNN on a different image database, although the proposed method provides no guarantee of test completeness.

DeepXplore \cite{S8} first introduced neuron coverage as a testing metric for DNNs, and then used multiple different DNNs with similar functionality to identify erroneous corner cases. Compared to \cite{S8}, DeepTest \cite{S10} and DLFuzz \cite{S53} aimed at maximizing the neuron coverage without requiring multiple DNNs. The study \cite{S10} employed metamorphic relations to identify erroneous behaviors. The study \cite{S53} proposed a differential fuzzing testing framework to generate adversarial inputs. However, methods proposed in \cite{S8, S10, S53} cannot guarantee the generation of test cases that can precisely reflect real-world cases (e.g., driving scenes in various weather conditions when taking a DNN-based autonomous driving system). DeepRoad \cite{S52} employed Generative Adversarial Network (GAN) based techniques and metamorphic testing to synthesize diverse real driving scenes, and to test inconsistent behaviors in DNN-based autonomous driving systems. In contrast to earlier works, DeepGauge \cite{S51} argued that the testing criteria for traditional software  are no longer applicable for DNNs. Ma et al. \cite{S51} proposed neuron-level and layer-level coverage criteria for testing DNNs and for measuring the testing quality.
\subsubsection{CA4: Assuring safety property of NN-based SCCPSs}
\label{SSS:6}
Formal verification can provide a mathematical proof that a system satisfies some desired safety properties (e.g., the system should always stay within some allowed region, namely a safe region). Formal verification usually presents NNs as models and then apply a model checker, such as Boolean satisfiability (SAT) solvers (e.g., Chaff \cite{RN8969}, SATO \cite{RN8970}, GRASP \cite{RN8968}) to verify the safety property. Pulina et al. \cite{S27} developed NeVer (“Ne”ural networks “Ver”ifier), which solves Boolean combinations of linear arithmetic constraints, to verify safe regions of MLPs. Through adopting an abstraction-refinement mechanism, NeVer can verify real-world MLPs automatically. As an extended experiment analysis of results of \cite{S27}, \cite{S6} compared the performance (e.g., competition-style and scalability) of state-of-the-art Satisfiability Modulo Theories (SMT) solvers \cite{RN8967}, and demonstrated that scalability and fine-grained abstractions remain  challenges for realistic size networks. The studies \cite{S17} and \cite{S35} verified the \textit{``feed-forward NNs with piece-wise linear activation functions''} by encoding verification problems into solving a linear approximation exploring network behavior in a SMT solver.

The next generation of collision avoidance systems for unmanned aircrafts (ACAS Xu) adopted DNNs to compress large score table \cite{julian2016policy}. Julian et al. \cite{S33} explored the performance of ACAS Xu by measuring a set of safety and performance metrics. A simulation in study \cite{S33} shows that the system based on DNNs performed as correctly as the original large score table but with better performance. Reluplex \cite{S35}  had successfully been used to verify the safety property of a  DNN for the prototype of ACAS Xu. Although the outcomes of Reluplex \cite{S35} are limited to verifying the correctness of NNs with specific type of activation functions (i.e., ReLUs and max-pooling layers), the study sheds a light on which types of NN architectures are easier to verify, and thus paves the way for verifying real-world DNN-based controllers. 

The method proposed in studies \cite{S45} and \cite{S46} verified that  Binarized Neural Networks (BNNs) are efficient and scalable to moderate-sized BNNs. Study \cite{S45} represented BNNs as boolean formulas, and then verified the robustness of BNNs against adversarial perturbations. In study \cite{S46}, BNNs and their input-output specifications were transferred into equivalence hardware circuits. The equivalence hardware circuits consist of a BNN structure module and a BNN property module. The authors of \cite{S46} then applied a SAT solver to verify the properties (e.g., “simultaneously classify an image as a priority road sign and as a stop sign with high confidence”) of the BNN in order to identify the risk behavior of the BNN. 

 When verifying a SCCS, one of the fundamental concerns is to make sure that the SCCS will never violate a safety property. An example of a safety property is that the system should never reach an unsafe region. The main ideas of studies under this sub-theme are to calculate the output reachable set of MLPs, such as in studies \cite{S31} and \cite{S34}, or  DNNs in study \cite{S30}, to verify if unsafe regions will be reached. Xiang et al. \cite{S34} proposed a layer-by-layer approach to compute the output reachable set assisted by polyhedron computation tools. The safety verification of a ReLU MLP is turned into checking if a non-empty intersection exists between the output reachable set and the unsafe regions. In a later work of Xiang et al. \cite{S31}, they introduced maximum sensitivity to perform a simulation-based reachable set estimation with few restrictions on the activation functions. By combining local search and linear programming problems, Dutta et al. \cite{S30} developed an output bound searching approach for DNNs with ReLU activation functions, which is implemented in a tool called SHERLOCK to check whether the unsafe region is reached. Study \cite{S36} focused on the safety verification of image classification decisions. In \cite{S36}, Huang et al. employed discretization to enable a finite exhaustive search for adversarial misclassifications. If no misclassifications are found in all layers after the exhaustive search, the NN is regarded as safe. 
 
 The idea of \cite{S50} was to formulate the formal verification of temporal logic properties of a CPS with Machine Learning (ML) components as the falsification problem (finding a counterexample that does not satisfy system specification). The study \cite{S50} adopted an ML analyzer to abstract the feature space of ML components (which  approximately represents  the ML classifiers). The identified misclassifying features are then used to drive the process of falsification. The introduction of the ML analyzer narrowed down the searching space for counterexamples and established a connection between the ML component and the rest of the system.
 
 Another direction to make sure the system will not violate safety properties is to use run-time monitoring. The study \cite{S56} envisioned an approach named WISEML, which combines reinforcement learning and run-time monitoring technique, to detect invariants violations. The purpose of this work was to create a safety envelope around the NN-based SCCPSs.

\subsubsection{CA5: Improving interpretability of NNs}
\label{SSS:7}
NNs have proved to be effective ways to generalize the relationship between inputs and outputs. As the models of NNs are learned from training data sets without human intervention, the relationship between the inputs and outputs of NNs is like a black box. Due to the black-box nature of NNs, it is difficult for people to understand and explain how an NN works. Studies under this theme focus on facilitating the understanding on how NNs generate outputs from inputs. Studies in this theme can be classified into the following three sub-themes, which can be overlapped. However, this can be a way to capture the different motivations for the interpretability of NNs.

\textbf{Studies focusing on understanding how a specific decision is made.} This line of work mainly focuses on providing explanations for individual predictions (also defined as local interpretability). One study is called Local Interpretable Model-agnostic Explanations (LIME) \cite{S48}. LIME can approximate the original NN model locally to provide an explanation for a specific prediction of interest. The problem of LIME is that it assumes the local linearity of the classification boundary, which is not true for most complex NNs. The creators of LIME later extended their work by introducing high-precision rules (i.e., if-then rules), which they called \textit{anchors} \cite{S58}. The study \cite{S57} developed an explanation system named LEMNA for security applications and Recurrent Neural Networks (RNNs). LEMNA can locally approximate  a non-linear classification boundary and handle feature dependency problems and therefore is able to provide a high fidelity explanation. 

In the case of an image classifier, it is also common to use gradient measurements to estimate the importance value of each pixel for the final classification. DeepLIFT \cite{S74}, Integrated Gradients \cite{S59}, and more recently, SmoothGrad \cite{S79} fall into this category. The study \cite{S83} proposed a unified framework, SHapley Additive exPlanations (SHAP), by integrating six existing methods (LIME \cite{S13}, DeepLIFT \cite{S74}, Layer-Wise Relevance Propagation, Shapley regression values, Shapley sampling values, and Quantitative Input Influence) to measure feature importance.

Several approaches attempted to decompose the classification decision (output) into the contributions of individual components of an input based on specific local decomposition rules (i.e., Pixel-Wise decomposition \cite{S60,S75}, and deep Taylor decomposition \cite{S64}). 

Szegedy et al. \cite{S24} investigated the semantic meaning of individual units and the stability of DNNs while small perturbations were added to the input. They pointed out that the individual neurons did not contain the semantic information, while the entire space of activations does. They also experimentally proved that the same small perturbation of input can cause different DNN models (e.g., trained with different hyperparameters) to generate wrong predictions. 

There are several methods for improving local explanations for NN models compared to the above-mentioned approaches. The study \cite{S72} argued that explanation approaches for NN models should provide sound theoretical support. \citet{S77} presented their idea as \textit{``Right for the right reasons,''} which means that the output of NN models should be right with the right explanation. In \citet{S77}, incorrect explanations for particular inputs can be identified, and NN models can be guided to learn alternate explanations. Both \cite{S72, S76} made efforts on real-time explanations since their approaches can generate accurate explanations quickly enough.

\textbf{Studies focusing on facilitating understanding of the internal logic of NNs.} Studies in this sub-theme are also known as global interpretability. To help interpret how NN models work, model distillation is used in \cite{S13}, \cite{S62}, \cite{S63}, and \cite{S71}. The initial intention of distillation was to reduce the computational cost. For example, \citet{S63} distilled a collection of DNN models into a single model to facilitate deployment. The knowledge distilled from NN models has later been applied for interpretability. Some studies compressed information (e.g., decision rules) from deep learning models into transparent models such as decision trees \cite{S13,S67} and gradient boosting trees \cite{S62} to mimic the performance of models. Others tended to explain the inner mechanisms of NN models through analyzing the feature space. Study \cite{S71} distilled the relationship between input features and model predictions (outputs of the model) as a feature shape to evaluate the feature contribution to the model. 

Another attempt to produce global interpretability is to reveal the features learned by each neuron. For example, in \cite{S80}, the authors leveraged deep generator networks to synthesized the input (i.e., image) that highly activates a neuron. \citet{S66} adopted an attentive encoder-decoder network to learn interpretable features, and then proposed an algorithm called \textit{prediction difference maximization} to interpret the features learned by each neuron.

One interesting work \cite{S78} used an additional NN module that is fit for relational reasoning to reason the relations between the input and response of the NN models. There is also another promising line of work (e.g., \cite{S65}, \cite{S73}) that combined local and global interpretability  to explain NN models.

\textbf{Studies focusing on visualizing internal layers of NNs to help identify errors in NNs.} In study\cite{S47}, activities, such as the operation of the classifier and the function of intermesdiate feature layers within the CNN model, were visualized by using a multi-layered deconvolutional network (named DeconvNet). These visualizations are useful to interpret model problems. Unlike \cite{S47}, which visually depicted neurons in a convolutional layer, the study \cite{S61} visualized neurons in a fully connected layer. \citet{S70} proposed \textit{Class Activation Mapping (CAM)} for CNNs to visualize the discriminative object parts on any given image. \citet{S68} highlighted the most responsible part of an image for a decision by perturbing meaningful images. DarkSight \cite{S69} combined the ideas of model distillation and visualization to visualize the prediction of an NN model.  \citet{S81} built a \textit{TreeView} representation via feature-space partitioning to interpret the prediction of an NN. Mahendran et al. \cite{S82} reconstructed semantic information (images) in each layer of CNNs by using information from the image representation.  

\subsection{RQ3. What are the limitations of current research with respect to testing and verifying NN-based SCCSs?}
\label{SS:3}
Analyzing failure modes and how the system reacts to failures are crucial parts of the safety analysis, especially in safety-critical domains. When testing and verifying the safety of NN-based SCCPSs, we need to rethink how to perform failure mode and effect analysis, how to analyze inter-dependencies between sub-systems of SCCPSs, and how to analyze the resilience of the system. We need to ensure that even if some of the system's hardware or software do not behave as expected, the system can sense the risk, avoid the risk before the incident, and mitigate the risk effectively when an incident happens. Looking into T\&V activities through software development, the ideal situation is that we would find appropriate T\&V methods to verify whether the design and implementation are consistent with the requirements, construct complete test criteria and test oracle, and generate test data and test any objects (such as code modules, data structures) that are necessary for the correct development of software \cite{adrion1982validation}. Unfortunately, the fact is that complete T\&V is hard to guarantee. In order to investigate the gap between industry needs for T\&V of NN-based SCCPS and state-of-the-art T\&V methods, we performed a mapping of identified approaches to the relevant standard.
\subsubsection{Mapping of reviewed approaches to the software safety lifecycles in IEC 61508}
\label{SSS:1}
An increased interest in the application of NNs within safety-critical domains has encouraged research in the area of T\&V of NN-based SCCSs. Research institutions and industry T\&V practitioners are working on different aspects of this problem. However, we have not found strong connections between those potentially useful methods for T\&V of NNs and relevant safety standards (such as IEC 61508 \cite{RN9010} and ISO 26262 \cite{ISO26262}). 

We hereby adopt IEC 61508 \cite{RN9010} as a reference standard to execute the mapping analysis since ISO 26262 \cite{ISO26262} is the adaptation of IEC 61508 \cite{RN9010}. We found that the major T\&V activities listed in the software safety lifecycles of IEC 61508-3 (including evaluation of software architecture design, software module testing and integration, programmable electronics integration, and software verification) are still valid when conducting T\&V for NN-based SCCSs. But for most of them, new techniques/measures for supporting the T\&V of NN-based software are demanded. Therefore, we decided to employ safety integrity properties (which are explained in IEC 61508-3 Annex C and Annex F of IEC 61508-7) as indicators to justify to what extent these desirable properties have been achieved by the state-of-the-art methods for T\&V of NN-based SCCSs. The detailed mapping information can be found in Table \ref{table:map}. 

\begin{footnotesize}
\setlength{\tabcolsep}{3pt}
\begin{longtable}{p{2cm}p{2cm}p{2cm}p{1cm}p{4cm}}
\caption{A mapping of reviewed approaches to IEC 61508 safety lifecycle} \label{table:map} \\
\toprule
\multicolumn{1}{l}{\textbf{Phase}} & \textbf{Property} & \multicolumn{1}{l}{\begin{tabular}[c]{@{}l@{}}\textbf{Relevant primary} \\ \textbf{studies}\end{tabular}} & \rotatebox{90}{\textbf{Category}} & \textbf{Remaining challenges} \\ \hline
\endfirsthead
\multicolumn{3}{c}%
{{\bfseries \tablename\ \thetable{} -- continued from the previous page}} \\
\toprule
\multicolumn{1}{l}{\textbf{Phase}} & \textbf{Property} & \multicolumn{1}{l}{\begin{tabular}[c]{@{}l@{}}\textbf{Relevant primary} \\ \textbf{studies}\end{tabular}} & \rotatebox{90}{\textbf{Category}} & \textbf{Remaining challenges} \\ \hline
\endhead
\hline 
\endfoot
Software  & Completeness & None & & N/A \\\cline{2-5}
 \multicolumn{1}{l}{\begin{tabular}[c]{@{}l@{}}architecture \\ design  \end{tabular}} & Correctness & \cite{S33} & CA4 & Training process of NN-based algorithm is time-consuming. \\\cline{2-5}
 & Freedom from intrinsic faults & \multicolumn{1}{l}{\begin{tabular}[c]{@{}l@{}}\cite{S15,S26,S32,S40,S19},\\ \cite{S29} - \cite{S42}\end{tabular}} & CA1 & \ding{182} Limited to specific model classes, or tasks (e.g.,  image classifier), or  small size NNs \cite{S26}; \ding{183} Not immune to adversarial adaptation \cite{S19}; \ding{184} Lack of understanding on how system can be free from different kinds of attacks other than adversarial examples. \\\cline{2-5}
 & \multicolumn{1}{l}{\begin{tabular}[c]{@{}l@{}}Understand- \\ ability\end{tabular}} & \cite{S24} - \cite{S82} & CA5 & \ding{182} Limited to specific model classes, or tasks (e.g., image classifier), or small size NN models \cite{S13}; \ding{183} Not able to provide real-time explanations; \ding{184} Lack of evaluation method for the explanation of NNs. \\\cline{2-5}
 & Verifiable and testable design & \cite{S5} & CA3 & \ding{182} Lack of integrated computer- aided toolchains to support the verification activities; \ding{183} Limited to specific models, tasks or NN size.\\
 &  & \cite{S17} & CA4 & \ding{182} Limited to specific NN architectures (i.e., piece-wise linear activation functions), need better understanding of NN architectures; \ding{183} Trade-off  between efficient verification and linear approximation of the NN behavior is not studied sufficiently. \\\cline{2-5}
& \multicolumn{1}{l}{\begin{tabular}[c]{@{}l@{}}Fault \\ tolerance\end{tabular}} & \cite{S2,S4,S28,S55,S22} & CA2 & \ding{182} Decouple the fault tolerance from the classification performance \cite{S4}; \ding{183} Lack of studies on unexpected environmental failures. \\\cline{2-5}
 & Defense against common cause failure & None &  & N/A  \\\hline
\multicolumn{1}{l}{\begin{tabular}[c]{@{}l@{}}Software module \\ testing and \\ integration\end{tabular}} & Completeness & \cite{S37,S42} & CA1 & Lack of comprehensive criteria to evaluate testing adequacy. \\
  &  & \cite{S8} - \cite{S53} & CA3 & Low fidelity of testing cases compared with real-world cases \cite{S10}. \\\cline{2-5}
 & Correctness & \multicolumn{1}{l}{\begin{tabular}[c]{@{}l@{}}\cite{S7,S23,S37,S3,S9}\\\cite{S11,S20}\end{tabular}} & CA1 & \ding{182} Vulnerable to the variation of adversarial examples; \ding{183} Limited to specific NN model classes or tasks. \\
 &  & \cite{S18} & CA2 & Insufficient validation of  input raw data. \\\cline{2-5}
 & Repeatability & \cite{S5,S8,S10} & CA3 & Testing cases  generated by automated tools may be biased. \\\cline{2-5}
 & Precisely defined testing configuration & None &  & N/A \\\hline
\multicolumn{1}{l}{\begin{tabular}[c]{@{}l@{}}Programm- \\able electronics \\ integration\\ (hardware \\ and software)\end{tabular}} & Completeness &None  &  & N/A \\\cline{2-5}
 & Correctness & \cite{S1,S44,S14,S43} & CA2 & Insufficient testing of hardware accelerator. \\\cline{2-5}
 & Repeatability & None &  & N/A \\\cline{2-5}
 & Precisely defined testing configuration & None &  & N/A \\\hline
\multicolumn{1}{l}{\begin{tabular}[c]{@{}l@{}}Software \\ verification\end{tabular}} & Completeness & \cite{S31,S34} & CA4 & \ding{182} Limited to specific NN models; \ding{183} Lack of scalability. \\\cline{2-5}
 & Correctness & \cite{S54} & CA2 & \ding{182} Automatic generation of complete testing scenarios sets. \\
 &  & \multicolumn{1}{l}{\begin{tabular}[c]{@{}l@{}}\cite{S6,S27,S30,S35,S36}\\ \cite{S45} - \cite{S50}\end{tabular}} & CA4 & \ding{182} Scalability and computational performance need to improve; \ding{183} SMT encoding for large-scale NN model; \ding{184} Lack of model-agnostic verification methods; \ding{185} Automatic generation of feature space abstractions \cite{S50}. \\\cline{2-5}
 & Repeatability & None &  & N/A  \\\cline{2-5}
 & Precisely defined testing configuration & None  &  & N/A\\\hline
\end{longtable}
\end{footnotesize}

In Table \ref{table:map}, we mapped existing T\&V methods for NN-based SCCSs (column 3 and column 4) into relevant properties (column 2) of four major T\&V phases (column 1) in the software safety lifecycles of IEC 61508-3. For column 5 in Table \ref{table:map}, we summarized the remaining challenges in testing and verifying NN-based SCCSs based on reviewed papers. The overviews of these remaining challenges can potentially inspire researchers to look for a focus in the future.  

\subsubsection{Limitations and suggestions for testing and verifying NN-based SCCSs}
\label{SSS:2}
In Table \ref{table:map}, we show the limitations and gaps of state-of-the-art T\&V approaches for NN-based SCCSs. In this section, we will take two T\&V phases (evaluation of software architecture design and software module testing and integration) as examples to provide detailed analysis of identified limitations and corresponding suggestions on the basis of required safety integrity properties. For the other two T\&V phases (programmable electronics integration and software verification), only summaries of limitations and suggestions will be presented to avoid duplication.

\paragraph{Evaluation of software architecture design}
The top three properties that have been addressed are: \textit{simplicity and understandability} (31 papers), \textit{freedom from intrinsic design faults} (10 papers), and \textit{fault tolerance} (5 papers). \textit{Correctness with respect to software safety requirements specification} (1 paper) and \textit{verifiable and testable design} have drawn little attention (2 papers) for reviewed studies. There are two properties, i.e., \textit{completeness with respect to software safety requirements specification} and \textit{Defense against common cause failure from external events}, which have not been addressed in reviewed papers. 
\subparagraph{Completeness with respect to software safety requirements specification}
No study contributes to the achievement of completeness, which requires the architecture design to be able to address all the safety needs and constraints. The achievement of completeness depends on the achievement of other properties, such as fully understanding the behavior of NN models. The design and deployment of NN-based SCCSs are in its infancy stage. When NN-based SCCS design becomes more practical, more studies may address this topic. 
\subparagraph{Correctness with respect to software safety requirements specification}
To achieve correctness, software architecture design needs to respond to the specified software safety requirements appropriately. Study \cite{S33} reported their successful design of a DNN-based compression algorithm for aircraft collision avoidance systems. Even though they demonstrated that the DNN-based algorithm preserves the required safety performance, the training process is still time-consuming.
\subparagraph{Freedom from intrinsic design faults}
Intrinsic design faults can be interpreted as failures derived from the design itself. State-of-the-art NNs have proved to be vulnerable to adversarial perturbations due to some intriguing properties of NNs \cite{S15}. Most of the studies in this category were aimed at understanding, detecting, and mitigating adversarial examples. Study \cite{S36} reported that their approach could generalize well on several state-of-the-art NNs to find adversarial examples successfully. However, the verification process of founded features is time-consuming, especially for larger images. In this sense, the scalability and computational performance of adversarial robustness are expected to be addressed in the future. In addition, adversarial robustness does not imply that the NN model is truly free from intrinsic design faults. How to assure freedom from interferences (e.g., signal-noise ratio degradation) other than adversarial perturbations is a research gap that needs to be filled.
\subparagraph{Understandability}
 This property can be interpreted as the predictability of system behavior, even in erroneous and failure situations. In this category, studies focusing on providing  explanations  for  individual prediction (e.g., \cite{S24}) and on visualizing internal layers of NN (e.g., \cite{S47,S48,S57}) are not meaningful for safety assurance. Studies focusing on facilitating understanding of the internal logic of NNs (such as presenting NNs as decision trees \cite{S13}) could be a solution to improve the understandability of NN-based architecture design. However, this line of work is rare, and most methods are only applied to small-scale DNNs with image input, or specific NN models. Besides, assuming the explanation of NN is available, confirming the correctness of the explanation is still a challenge. Interpretability of NNs is undoubtedly a crucial need in safety-critical applications. Methods in this line should capable of explaining different types of sensor data (e.g., image, text, and point data) and both local and global decisions.
\subparagraph{Verifiable and testable design}
The evaluation metrics of verifiable and testable design may be derived from modularity, simplicity, provability, and so on. We observed that existing verifiable and testable designs are limited to specific NN architectures (e.g., \cite{S17}) or specific tasks (e.g., \cite{S5}). There is no standard procedure for determining which type of NNs will be easier to verify. \citet{S17} argued that NNs that adopt piece-wise linear activation functions are easier to verify, but their method still need to face the conflict between efficient verification and accuracy of linear approximation for the NN behavior.
\subparagraph{Fault tolerance}
 Fault tolerance implies that the architecture design can assure the safe behavior of the software whenever internal or external errors occur. To achieve fault tolerance, features like failure detection and failure impact mitigation of both internal and external errors should be included in the design. Existing methods showed that unexpected environmental failures are hard to detect and mitigate. Besides, many of the proposed approaches in this category have not yet been evaluated in the real-world. Some studies formulated approximated computational models to represent real-world systems (e.g., \cite{S2}). The study \cite{S22} did not use any test oracle when executing system flight tests.  Some studies used simulation models to verify the performance of the original NN (e.g., \cite{S4}). They are not able to prove the fidelity of the model compared with the real-world system. 
\subparagraph{Defense against common cause failure from external events}
Software common cause failure is a type of concurrent failure of two or more modules in the software, which is caused by software design defects and triggered by external events such as time, unexpected input data, or hardware abnormalities \cite{international2009iaea}. Many safety critical systems adopt redundant architectures (meaning two or more independent subsystems have identical functions to back-up each other) to prevent a single point of failure. However, redundant architectures are vulnerable considering common cause failure. In the context of NN-based SCCSs, it is common to employ multiple NNs with similar architectures in order to improve the accuracy of prediction. If a common cause failure occurs in this kind of software design, the prediction might be totally wrong, and thus the control software might make the wrong decision.
DeepXplore, reported in \cite{S8}, used more than two different DNNs with the same functionality to automatically generate a test case. If all the DNNs in DeepXplore are affected by common cause failure, such as if a sensor failure causes all the DNNs to make the same misclassification, then it will not be able to generate the corresponding test case. No method is found in reviewed papers that can identify common cause failure modes and defend against such failures. In order to effectively defend against common cause failure, designers need to inspect the completeness and correctness of the safety requirements specification, trace the implementation of the safety requirements specification, and make a thorough T\&V plan to reveal the common cause failure modes in the early stage.
\paragraph{Software module testing and integration}
The top two properties that have been addressed are: \textit{completeness of testing and integration with respect to the design specifications} (9 papers) and \textit{correctness of testing and integration with respect to the design specifications} (8 papers). \textit{Repeatability} has drawn little attention (3 papers) from the reviewed studies. There is one property, \textit{precisely defined testing configuration}, which has not been addressed in the reviewed papers. This property aims to evaluate the precision of T\&V procedures, which is not in the scope of our selected papers. Therefore, we will not give more explanation on this property. 
\subparagraph{Completeness of testing and integration with respect to the design specifications}
We observed some efforts that tried to find a systematic way to generate testing cases (e.g., \cite{S10,S52}) to measure testing quality (e.g., \cite{S51}) or to connect different T\&V stages in the development of SCCSs (e.g., \cite{S25}). As analyzed in Section \ref{SS:2}, we can infer that an NN-based control software is instinctually different in design workflow and software development compared to the design of traditional control software. We suggest that the testing criteria should thoroughly align with the software design. To be more specific, the instinctive features of NN-based softwares (e.g., NN model's architectural details and the working mechanism of NNs) should be carefully considered when setting the testing criteria. That is testing criteria should be defined comprehensively and explicitly under the consideration of not only test case coverage but also the robustness of NN-based system performance (for instance, test how an NN will respond when input data change slightly) and the features of training data sets, such as the data density issue mentioned in  \cite{RN8986}.
\subparagraph{Correctness of testing and integration with respect to the design specifications}
Several studies (e.g., \cite{S7,S3,S9}) reported that their methods are vulnerable to the variation of adversarial examples. Another common limitation is that most methods are model-specific, meaning that they can only apply to a single type or class of NN model. To achieve correctness of testing and integration, the module testing task should be completed, which means the testing should cover both NN models and external input. However, few studies focused on the validation of input data. One study \cite{S18} identified that sufficient validation of input raw data remains a challenge.
\subparagraph{Repeatability}
The complexity and un-interpretable feature of NNs make manual testing almost infeasible. In order to be able to generate consistent results from testing repeatedly, some studies were dedicated to achieving automatic test execution or even automatic test generation. We found three papers (i.e., \cite{S5, S8, S10}) addressing automatic test generation. However, generating test cases automatically is still a challenge. For instance, studies \cite{S8,S10} claimed that the test cases generated by an automated testing tool may not cover all real-world cases. 

\paragraph{Programmable electronics integration}
 The major limitation of this line of work is insufficient testing for hardware accelerators. NN-based SCCPSs requires typically high-performance computing systems, such as Graphics Processing Units (GPUs). Some industry participants have provided specialized hardware accelerators to accelerate NN-based computations. For example, Google deployed a DNN accelerator (called Tensor Processing Unit) in its data centers for DNN applications \cite{RN8973}. NVIDIA introduced an automotive supercomputing platform named DRIVE PX 2 \cite{RN577}, which now has been used by over 370 companies and research institutions in the automotive industry \cite{partners}. However, little research effort has been put into the T\&V of the reliability of using hardware accelerators for NN applications. We found seven studies (i.e. \cite{S1, S2, S4, S14, S44, S18, S43}) addressing the evaluation of the error resilience of hardware accelerators. However, the testing is limited to specific type errors (e.g., radiation-induced soft errors, which are presented in \cite{S1,S14,S43}). The mitigation method proposed in \cite{S14} (called ABFT: Algorithm-Based Fault Tolerance) can only protect portions of the accelerator (e.g., sgemm kernels, which is one kind of matrix multiplication kernels). The study \cite{S18} identified errors made by single frame object detectors, but the result showed that the method is not capable of detecting all mistakes. The studies \cite{S1,S43} investigated the propagation characteristic of soft errors in the DNN system, but they used a DNN simulator instead of a real DNN accelerator for fault injection.
\paragraph{Software verification}
In general, there is a lack of a comprehensive and standardized framework for verifying the safety of NN-based SCCSs. Formal verification procedures are highly demanding. The common limitation of formal verification approaches is the scalability issues. Most proposed methods are limited to a specific NN structure and size (e.g., \cite{S17,S27,S35,S45,S46}). The study \cite{S27}  reported that their approaches can only verify small-scale systems  (i.e., the layer of NN is 3 and the maximum amount of input neurons is 64). One approach reported in \cite{S45} can verify medium size NNs. The verification of large-scale NNs is still a challenge. Another limitation is that proposed approaches are not robust to NN variations. For example, verification methods in studies \cite{S17,S35} are only adapted to specific network types and sizes.
\section{Discussion}
\label{S:5} 
In this section, we first discuss industry practices for T\&V of NN-based SCCPSs. Then, we compare this SLR with related works. At the end of this section, we present the threats to the validity of our study.
\subsection{Industry practice}
Our findings on the research questions (RQ1 to RQ3) mainly reflected the academic efforts addressing T\&V of NN-based SCCPSs. NN-based applications have drawn a lot of attention from industry practitioners. Taking the automotive industry as an example, several car makers (e.g., GM, BMW, and Tesla) and some high technology companies (e.g., Waymo and Baidu) are leading the revolution in autonomous driving safety.
\subsubsection{Safety of the intended functionality}
At the beginning of this year, ISO/PAS 21448:2019 \cite{RN8974} was published. It listed recommended methods for deriving verification and validation activities (See ISO/PAS 21448:2019 Table 4). In Table \ref{table:SOTIF}, we highlighted six of the recommended methods, which shared similar verification interests with existing academic efforts.
\begin{table}[ht]
\caption{Shared verification interests of ISO/PAS 21448 and academic efforts}
\label{table:SOTIF}
\resizebox{\textwidth}{!}{%
\begin{tabular}{l||l}
\toprule
\textbf{ISO/PAS 21448} & \textbf{Academic efforts} \\
\hline
Analysis of triggering events & CA1: Assuring robustness of NNs \\
Analysis of sensors design and their known potential limitations & CA2: Improving failure resilience of NNs \\
Analysis of environmental conditions and operational use cases & CA3: Measuring and ensuring test completeness \\
Analysis of boundary values & CA4:  Assuring safety property of NN-based SCCPSs \\
Analysis of algorithms and their decision paths & CA5:  Improving interpretability of NNs \\
Analysis of system architecture & CA1-CA5\\
\bottomrule
\end{tabular}
}
\end{table}
\subsubsection{Safety reports}
In 2018, three companies (Waymo, General Motor, and Baidu Apollo) published their annual safety reports. As a pioneer in the development of self-driving cars, Waymo proposed the ``\textit{Safety by Design}'' \cite{waymo} approach, which entails the processes and techniques they used to face safety challenges of a level 4 autonomous car on the road. For the cybersecurity consideration, Waymo adopted Google's security framework \cite{google} as the foundation. After that, General Motor (GM) released their safety report \cite{GM} for Cruise AV (also level 4). GM's safety process combined conventional system validation (such as vehicle performance tests, fault injection testing, intrusive testing, and simulation-based software validation) with SOTIF validation through iterative design. Baidu adopted the Responsibility-Sensitive Safety model \cite{2017arXiv170806374S} proposed by Mobileye \cite{mobileye} (an Intel company) to design the safety process for the Apollo Pilot for a passenger car (level 3). 

In addition, we noticed that Tesla started releasing quarterly safety data  since October 2018 \cite{Tesla}. It seemed that Tesla has a completely different approach to self-driving cars than other companies. According to TESLA NEWS \cite{Autopilot}, AutoPilot will rely for its self-driving function on cameras, not on LIDAR; the AutoPilot software is trained online (which means that the NN keeps learning and evolving during operation).  The Autopilot’s safety features are continuously evolved and enhanced through understanding real-world driving data from every Tesla. 

Referring to these safety reports of existing autonomous cars, we should be aware that when testing DNN-based control software (the core part of autonomous vehicles), black-box system level testing (by observing inputs and its corresponding outputs, e.g., closed course testing and real-world driving) is still the leading method. More systematic T\&V criteria and approaches are needed for more complete and reliable testing results. 

\subsection{Comparison with related work}
\subsubsection{Verification and validation of NNs}

\citet{taylor2003verification} conducted a survey on the Verification and Validation (V\&V) of NNs used in safety-critical domains in 2003. Study \cite{taylor2003verification} is the closest work we found, although they did not adopt an SLR approach. Our study covered new studies from 2011-2018. The authors of \cite{taylor2003verification} also made a classification of  methods for the V\&V of NNs. They grouped the methods into five traditional V\&V technique categories, namely, automated testing and testing data generation methods, run-time monitoring, formal methods, cross validation, and visualization. In contrast to \cite{taylor2003verification}, our study adopted a thematic analysis approach \cite{RN8963} and identified five themes based on the research goals of the selected studies. We thought it was better to classify the proposed T\&V methods of NNs based on their aims rather than on the traditional technique categories since many traditional V\&V techniques are no longer effective for verifying NNs in many cases. New methods and tools should be explored and developed without being limited by the traditional V\&V categories. Another difference is our study specialized more in the T\&V of modern NNs, such as MLP and DNN, whereas the study \cite{taylor2003verification} provided more in-depth analysis of V\&V methodologies for NNs used in flight control system, such as Pre-Trained Neural Network (PTNN) and Online Learning Neural Network (OLNN). Our study and \cite{taylor2003verification} have some common findings. For example, one category, named \textit{Visualization} in \cite{taylor2003verification}, falls into our category CA5 Improving interpretability of NNs.

\subsubsection{Surveys of security, safety, and productivity for Deep Learning (DL) systems development}

    Hains et al. \cite{hains2018towards} surveyed existing work on ``\textit{attacks against DL systems; testing, training, and monitoring DL systems for safety; and the verification of DL systems.}''  Our study and \cite{hains2018towards} shared a similar motivation. The critical difference between our SLR and \cite{hains2018towards} are threefold: 1) We conducted our literature review on 83 selected papers based on specific SLR guidelines, while they used an ad hoc literature review (ALR) approach and reviewed only 21 primary papers. 2) They only focused on DL systems, whereas our scope covered modern NN-based software systems, which embodies DL-based software systems. 3) They inferred that formal methods and automation verification are the two promising research directions  based on the reviewed works. In contrast, we focused more on safety issues, and found more categories to be addressed for safety purposes. 
    
\subsubsection{Surveys of certification of AI technologies in automotive}

Falcini et al. \cite{RN8559,RN8895} reviewed the existing standards in the automotive industry and pointed out the related applicability issues of automotive software development standards to deep learning. Although our SLR takes the automotive industry as an example, we are concerned with SCCPSs in general. This concern is reflected in the distribution of the selected papers (only 13 of the 83 selected papers are oriented to automotive CPSs).
    
\subsubsection{SLR of Explainable Artificial Intelligence (XAI)}

There are two very recent SLRs, \cite{RN9843} and \cite{RN8571}, on the interpretation of artificial intelligence. Both \cite{RN9843} and \cite{RN8571} employed similar commonly accepted guidelines to conduct their SLRs. The fundamental difference between our study and \cite{RN8571,RN9843} is the scope. \cite{RN9843} reviewed 381 papers on existing XAI approaches from interdisciplinary perspectives. As reported in \cite{RN8571}, the scope of their SLR is visualization and visual analytics for deep learning. The study \cite{RN8571} focused on studies that adopted visual analytics to explain NN decisions. Our study has a more comprehensive coverage of T\&V approaches that were employed to not only interpret NN behaviors but also to assure the robustness of NNs, to improve the failure resilience of NNs, to ensure test completeness, and to assure the safety property of NN-based SCCPSs. In a summary, our SLR tried to provide an overview of key aspects related to T\&V activities for NN-based SCCSs.

\subsection{Threats to validity}

In this section, we discuss some threats to the validity of our study.

\subsubsection{\textbf{Search strategy}} 
The most possible threat in this step is missing or excluding relevant papers. To mitigate this threat, we used six of the most relevant digital libraries to retrieve papers. Additionally, we employed two strategies to mitigate potential limitations in the search terms: 1) adopted an PIOC criteria to ensure the coverage of search terms; and 2) improved search terms iteratively. Further, we conducted an extensive snowballing process on references of the selected papers to identify related papers. The search keywords were cross-checked and agreed on by both authors. 

\subsubsection{\textbf{Study selection}} Researchers' subjective judgment could be a threat to the study selection. We strictly followed the pre-defined review protocol to mitigate this threat. For example, we started recording the inclusion and exclusion reasons from the 3rd stage. We validated the inclusion and exclusion criteria with two authors on the basis of the pilot search. Furthermore, the second author performed a cross-check of all selected papers. Any paper that raised doubts about its inclusion or exclusion decision was discussed between the first and second authors. For example, the \textit{``smart grid''} is included in the search term, but no relevant papers were found after the 3rd stage. Then, we conducted a snowballing search to identify papers that presented how to use NNs in smart grids. We found out that AI  is mainly used to solve the economically relevant problems \cite{RN8959} of the smart grid system (e.g., prediction of energy usage and efficient use of resources). AI is not involved in the safety-critical applications (e.g., decision making on optimal provision of power) of smart grids. Therefore, there were no relevant papers addressing safety analysis or testing/verification (refer to Inclusion criteria I2).

\subsubsection{\textbf{Data extraction}} The first author was responsible for designing the data extraction form and conducting the data extraction from selected papers. In order to avoid the first author's bias in data extraction, the two authors continuously discussed the data extraction issues. The extracted data were verified by the second author.

\subsubsection{\textbf{Data synthesis}} Data analysis outcomes could vary with different researchers. To reduce the subjective impact on data synthesis, besides strictly following the thematic synthesis steps \cite{RN8963}, the data synthesis was first agreed on by both authors. We disseminated our preliminary findings to two internal research groups at our university (i.e., the autonomous vehicle lab and autonomous ships lab) and presented at a Ph.D. seminar on IoT, Machine Learning, Security, and Privacy for comments and feedback. In summary, the audiences agreed with our research design and results, and they thought that the mapping of reviewed approaches to the IEC61508 is a valuable attempt. Several researchers working in formal verification and safety verification thought that safety cases would be a promising direction to address the challenges of T\&V of NN-based SCCSs. One suggestion is adding information about self-driving car simulators. Based on these comments and feedback, we revised our paper accordingly. 
\section{Conclusion and future work}
\label{S:6} 
In this paper, we have presented the results of a Systematic Literature Review (SLR) of existing approaches and practices on T\&V methods for neural-network-based safety critical control software (NN-based SCCS).  The motivation of this study was to provide an overview of the state-of-the-art T\&V of safety-critical NN-based SCCSs and to shed some light on potential research directions. Based on pre-defined inclusion and exclusion criteria, we selected 83 papers that were published between 2011 and 2018.  A systematic analysis and synthesis of the data extracted from the papers and comprehensive reviews of industry practices (e.g., technical reports, standards, and white papers) related to our RQs were performed. Results of the study show that:
\begin{enumerate}
\item The research on T\&V of NN-based SCCSs is gaining interest and attention from both software engineering and safety engineering researchers/practitioners according to the impressive upward trend in the number of papers on T\&V of NN-based SCCSs (See Fig. \ref{fig:yeartype}). Most of the reviewed papers (68/83, 81.9\%) have been published in the last three years.
\item The approaches and tools reported for the T\&V of NN-based control software have been applied to a wide variety of safety-critical domains, among which “automotive CPSs” has received the most attention. 
\item The approaches can be classified into five high-order themes, namely, assuring robustness of NNs, improving failure resilience of NNs, measuring and ensuring test completeness, assuring safety properties of NN-based SCCPSs, and improving interpretability of NNs.
\item The activities listed in the software safety lifecycles of IEC 61508-3 are still valid when conducting safety verification for NN-based control software. However, most of the activities need new techniques/measures to deal with the new characteristics of NNs.
\item Four safety integrity properties within the four major safety lifecycle phases, namely, correctness, completeness, freedom from intrinsic faults, and fault tolerance, have drawn the most attention from the research community. Little effort has been put on achieving repeatability. No reviewed study focused on precisely defined testing configuration and defense against common cause failure, which are extremely crucial for assuring the safety of a production-ready NN-based SCCS \cite{RN9878}. 
\item It is common to combine standard testing techniques with formal verification when testing and verifying large-scale, complex safety-critical software \cite{taylor2003verification,adrion1982validation}. As explained in section \ref{SS:3}, we found that an increasing concern of the reviewed works is the integration of different T\&V techniques in a systematic manner to gain assurance for the whole lifecycle of the NN-based control software. 
\end{enumerate}

This SLR is just a starting point in our studies to test and verify NN-based SCCPSs. In the future, we will focus on improving the interpretability of NNs. To be more specific, we plan to develop a method for explaining why an NN model is more (or less) robust than other models. It can guide software designers to design an NN model with an appropriate robustness level, which will greatly support safety by design.

\section*{Acknowledgments}
The authors would like to thank Weifeng Liu for commenting on and improving this paper. This work is supported by the Safety, autonomy, remote control and operations of industrial transport systems (SAREPTA) project, which is financed by the Norwegian Research Council with Grant No. 267860. This work is also supported by the Management of safety and security risks for cyber-physical systems project, which is
financed by the Norwegian University of Science and Technology and the Technical University of Denmark.



\bibliographystyle{model1-num-names}
\bibliography{SLR.bib}






\clearpage
\appendix
\section{Selected studies (sorted based on publication year)}
\begin{footnotesize}
\setlength{\tabcolsep}{3pt}
\label{my-label}
\begin{longtable}{lp{3cm}lp{6cm}p{4cm}}
\toprule
\multicolumn{1}{l}{\textbf{S\_ID}} & \textbf{Author(s)} & \multicolumn{1}{l}{\textbf{Year}} & \textbf{Title} & \textbf{Publication Venue} \\ \hline
\endfirsthead
\toprule
\multicolumn{1}{l}{\textbf{S\_ID}} & \textbf{Author(s)} & \multicolumn{1}{l}{\textbf{Year}} & \textbf{Title} & \textbf{Publication Venue} \\ \hline
\endhead
\hline
\endfoot
\cite{S27} & Pulina, L. and A. Tacchella & 2011 & NeVer: a tool for artificial neural networks verification & Annals of Mathematics and Artificial Intelligence \\
\cite{S6} & Pulina, L. and A. Tacchella & 2012 & Challenging SMT solvers to verify neural networks & AI Communications \\
\cite{S61} & Simonyan, K., A. Vedaldi and A. Zisserman & 2013 & Deep inside convolutional networks: Visualising image classification models and saliency maps & arXiv preprint \\
\cite{S24} & Szegedy, C., W. Zaremba, I. Sutskever, J. Bruna, D. Erhan, I. Goodfellow and R. Fergus & 2013 & Intriguing properties of neural networks & arXiv preprint \\
\cite{S15} & Goodfellow, I. J., J. Shlens and C. Szegedy & 2014 & Explaining and Harnessing Adversarial Examples & International Conference on Learning Representations (ICLR)\\
\cite{S40} & Gu, S. and L. Rigazio & 2014 & Towards deep neural network architectures robust to adversarial examples & International Conference on Learning Representations (ICLR) \\
\cite{S47} & Zeiler, M. D. and R. Fergus & 2014 & Visualizing and understanding convolutional networks & European conference on computer vision \\
\cite{S2} & Zhang, Q., T. Wang, Y. Tian, F. Yuan and Q. Xu & 2015 & ApproxANN: an approximate computing framework for artificial neural network & Design, Automation \& Test in Europe Conference \& Exhibition \\
\cite{S62} & Che, Z., S. Purushotham, R. Khemani and Y. Liu & 2015 & Distilling knowledge from deep networks with applications to healthcare domain & arXiv preprint \\
\cite{S63} & Hinton, G., O. Vinyals and J. Dean & 2015 & Distilling the knowledge in a neural network & arXiv preprint \\
\cite{S7} & Nguyen, A., J. Yosinski and J. Clune & 2015 & Deep neural networks are easily fooled: High confidence predictions for unrecognizable images & IEEE Conference on Computer Vision and Pattern Recognition (CVPR) \\
\cite{S75} & Bach, S., A. Binder, G. Montavon, F. Klauschen, K.-R. Müller and W. Samek & 2015 & On pixel-wise explanations for non-linear classifier decisions by layer-wise relevance propagation & PloS one \\
\cite{S38} & Scheibler, K., L. Winterer, R. Wimmer and B. Becker & 2015 & Towards Verification of Artificial Neural Networks & Workshop on Methods and Description Languages for Modeling and Verification of Circuits and Systems (MBMV) \\
\cite{S42} & Shaham, U., Y. Yamada and S. Negahban & 2015 & Understanding adversarial training: Increasing local stability of neural nets through robust optimization & arXiv preprint \\
\cite{S82} & Mahendran, A. and A. Vedaldi & 2015 & Understanding deep image representations by inverting them & IEEE conference on computer vision and pattern recognition \\
\cite{S60} & Bach, S., A. Binder, K.-R. Müller and W. Samek & 2016 & Controlling explanatory heatmap resolution and semantics via decomposition depth & IEEE International Conference on Image Processing (ICIP) \\
\cite{S12} & Papernot, N., P. McDaniel, X. Wu, S. Jha and A. Swami & 2016 & Distillation as a defense to adversarial perturbations against deep neural networks & IEEE Symposium on Security \& Privacy \\
\cite{S21} & Zheng, S., Y. Song, T. Leung and I. Goodfellow & 2016 & Improving the robustness of deep neural networks via stability training & IEEE conference on computer vision and pattern recognition \\
\cite{S22} & Daftry, S., S. Zeng, J. A. Bagnell and M. Hebert & 2016 & Introspective perception: Learning to predict failures in vision systems & IEEE/RSJ International Conference on Intelligent Robots and Systems (IROS) \\
\cite{S70} & Zhou, B., A. Khosla, A. Lapedriza, A. Oliva and A. Torralba & 2016 & Learning deep features for discriminative localization & IEEE conference on computer vision and pattern recognition \\
\cite{S26} & Bastani, O., Y. Ioannou, L. Lampropoulos, D. Vytiniotis, A. Nori and A. Criminisi & 2016 & Measuring neural net robustness with constraints & Advances in neural information processing systems \\
\cite{S74} & Shrikumar, A., P. Greenside, A. Shcherbina and A. Kundaje & 2016 & Not just a black box: Interpretable deep learning by propagating activation differences & arXiv Preprint \\
\cite{S33} & Julian, K. D., J. Lopez, J. S. Brush, M. P. Owen and M. J. Kochenderfer & 2016 & Policy compression for aircraft collision avoidance systems & IEEE/AIAA international conference on Digital Avionics Systems Conference (DASC)\\
\cite{S80} & Nguyen, A., A. Dosovitskiy, J. Yosinski, T. Brox and J. Clune & 2016 & Synthesizing the preferred inputs for neurons in neural networks via deep generator networks & Advances in Neural Information Processing Systems \\
\cite{S81} & Thiagarajan, J. J., B. Kailkhura, P. Sattigeri and K. N. Ramamurthy & 2016 & TreeView: Peeking into deep neural networks via feature-space partitioning & arXiv preprint \\
\cite{S44} & Li, G., K. Pattabiraman, C.-Y. Cher and P. Bose & 2016 & Understanding error propagation in GPGPU applications & International Conference on High Performance Computing, Networking, Storage and Analysis\\
\cite{S48} & Ribeiro, M. T., S. Singh and C. Guestrin & 2016 & Why should i trust you?: Explaining the predictions of any classifier & ACM SIGKDD International Conference on Knowledge Discovery and Data Mining \\
\cite{S59} & Sundararajan, M., A. Taly and Q. Yan & 2017 & Axiomatic attribution for deep networks & International Conference on Machine Learning \\
\cite{S5} & O'Kelly, M., H. Abbas and R. Mangharam & 2017 & Computer-aided design for safe autonomous vehicles & Resilience Week (RWS)\\
\cite{S50} & Tommaso DreossiAlexandre DonzSanjit A. Seshia & 2017 & Compositional Falsification of Cyber-Physical Systems with Machine Learning Components & NASA Formal Methods \\
\cite{S10} & Tian, Y., K. Pei, S. Jana and B. Ray & 2017 & DeepTest: Automated testing of deep-neural-network-driven autonomous cars & arXiv preprint \\
\cite{S11} & Reuben, F., R. R. Curtin, S. Saurabh and A. B. Gardner & 2017 & Detecting Adversarial Samples from Artifacts & arXiv preprint \\
\cite{S13} & Frosst, N. and G. Hinton & 2017 & Distilling a Neural Network Into a Soft Decision Tree & arXiv preprint \\
\cite{S8} & Pei, K., Y. Cao, J. Yang and S. Jana & 2017 & DeepXplore: Automated Whitebox Testing of Deep Learning Systems &  ACM Symposium on Operating Systems Principles (SOSP) \\
\cite{S9} & Gopinath, D., G. Katz, C. S. Pasareanu and C. Barrett & 2017 & Deepsafe: A data-driven approach for checking adversarial robustness in neural networks & arXiv preprint \\
\cite{S64} & Montavon, G., S. Lapuschkin, A. Binder, W. Samek and K.-R. Müller & 2017 & Explaining nonlinear classification decisions with deep Taylor decomposition & Pattern Recognition \\
\cite{S14} & Santos, F. F. d., L. Draghetti, L. Weigel, L. Carro, P. Navaux and P. Rech & 2017 & Evaluation and Mitigation of Soft-Errors in Neural Network-Based Object Detection in Three GPU Architectures & IEEE/IFIP International Conference on Dependable Systems and Networks Workshops (DSN-W) \\
\cite{S16} & Papernot, N. and P. McDaniel & 2017 & Extending defensive distillation & arXiv preprint \\
\cite{S17} & Ehlers, R. & 2017 & Formal verification of piece-wise linear feed-forward neural networks & International Symposium on Automated Technology for Verification and Analysis \\
\cite{S18} & Manikandasriram, S. R., C. Anderson, R. Vasudevan and M. Johnson-Roberson & 2017 & Failing to learn: autonomously identifying perception failures for self-driving cars & arXiv preprint \\
\cite{S19} & Xu, W., D. Evans and Y. Qi & 2017 & Feature squeezing: Detecting adversarial examples in deep neural networks & Network and Distributed Systems Security Symposium (NDSS) \\
\cite{S66} & Dong, Y., H. Su, J. Zhu and B. Zhang & 2017 & Improving interpretability of deep neural networks with semantic information & IEEE Conference on Computer Vision and Pattern Recognition \\
\cite{S67} & Bastani, O., C. Kim and H. Bastani & 2017 & Interpretability via model extraction & arXiv preprint \\
\cite{S68} & Fong, R. C. and A. Vedaldi & 2017 & Interpretable explanations of black boxes by meaningful perturbation & IEEE International Conference on Computer Vision \\
\cite{S23} & Melis, M., A. Demontis, B. Biggio, G. Brown, G. Fumera and F. Roli & 2017 & Is Deep Learning Safe for Robot Vision? Adversarial Examples Against the iCub Humanoid & IEEE International Conference on Computer Vision Workshops (ICCVW) \\
\cite{S25} & Vishnukumar, H. J., B. Butting, C. Muller and E. Sax & 2017 & Machine learning and deep neural network - artificial intelligence core for lab and real-world test and validation for ADAS and autonomous vehicles: AI for efficient and quality test and validation & Intelligent Systems Conference (IntelliSys)\\
\cite{S28} & Mhamdi, E. M. E., R. Guerraoui and S. Rouault & 2017 & On the Robustness of a Neural Network & IEEE Symposium on Reliable Distributed Systems (SRDS) \\
\cite{S29} & Metzen, J. H., T. Genewein, V. Fischer and B. Bischoff & 2017 & On detecting adversarial perturbations & International Conference on Learning Representations (ICLR) \\
\cite{S30} & Dutta, S., S. Jha, S. Sanakaranarayanan and A. Tiwari & 2017 & Output range analysis for deep neural networks & arXiv preprint \\
\cite{S32} & Cisse, M., P. Bojanowski, E. Grave, Y. Dauphin and N. Usunier & 2017 & Parseval networks: Improving robustness to adversarial examples & arXiv preprint \\
\cite{S34} & Xiang, W., H.-D. Tran and T. T. Johnson & 2017 & Reachable set computation and safety verification for neural networks with ReLU activations & arXiv preprint \\
\cite{S35} & Katz, G., C. Barrett, D. L. Dill, K. Julian and M. J. Kochenderfer & 2017 & Reluplex: An efficient SMT solver for verifying deep neural networks & International Conference on Computer Aided Verification (CAV)\\
\cite{S76} & Dabkowski, P. and Y. Gal & 2017 & Real time image saliency for black box classifiers & Advances in Neural Information Processing Systems (NIPS) \\
\cite{S77} & Ross, A. S., M. C. Hughes and F. Doshi-Velez & 2017 & Right for the right reasons: Training differentiable models by constraining their explanations & arXiv preprint \\
\cite{S4} & Vialatte, J.-C. and F. Leduc-Primeau & 2017 & A Study of Deep Learning Robustness Against Computation Failures & arXiv preprint \\
\cite{S78} & Santoro, A., D. Raposo, D. G. Barrett, M. Malinowski, R. Pascanu, P. Battaglia and T. Lillicrap & 2017 & A simple neural network module for relational reasoning & Advances in Neural Information Processing Systems (NIPS) \\
\cite{S36} & Huang, X. W., M. Kwiatkowska, S. Wang and M. Wu & 2017 & Safety Verification of Deep Neural Networks & International Conference on Computer Aided Verification\\
\cite{S79} & Smilkov, Daniel and Thorat, Nikhil and Kim, Been and Viégas, Fernanda and Wattenberg, Martin & 2017 & Smoothgrad: removing noise by adding noise & arXiv preprint \\
\cite{S37} & Carlini, N. and D. Wagner & 2017 & Towards Evaluating the Robustness of Neural Networks & IEEE Symposium on Security and Privacy (SP) \\
\cite{S41} & Katz, G., C. Barrett, D. L. Dill, K. Julian and M. J. Kochenderfer & 2017 & Towards proving the adversarial robustness of deep neural networks & arXiv Preprint \\
\cite{S43} & Li, G., S. K. S. Hari, M. Sullivan, T. Tsai, K. Pattabiraman, J. Emer and S. W. Keckler & 2017 & Understanding error propagation in deep learning neural network (DNN) accelerators and applications & International Conference for High Performance Computing, Networking, Storage and Analysis \\
\cite{S83} & Lundberg, S. M. and S.-I. Lee & 2017 & A unified approach to interpreting model predictions & Advances in Neural Information Processing Systems (NIPS) \\
\cite{S45} & Narodytska, N., S. P. Kasiviswanathan, L. Ryzhyk, M. Sagiv and T. Walsh & 2017 & Verifying properties of binarized deep neural networks & arXiv preprint \\
\cite{S49} & Raj, S., S. K. Jha, A. Ramanathan and L. L. Pullum & 2017 & Work-in-progress: testing autonomous cyber-physical systems using fuzzing features from convolutional neural networks & International Conference on Embedded Software (EMSOFT) \\
\cite{S1} & Schorn, C., A. Guntoro and G. Ascheid & 2018 & Accurate neuron resilience prediction for a flexible reliability management in neural network accelerators & Design, Automation \& Test in Europe Conference \& Exhibition (DATE) \\
\cite{S58} & Ribeiro, M. T., S. Singh and C. Guestrin & 2018 & Anchors: High-precision model-agnostic explanations & AAAI Conference on Artificial Intelligence \\
\cite{S51} & Ma, L., F. Juefei-Xu, F. Zhang, J. Sun, M. Xue, B. Li, C. Chen, T. Su, L. Li and Y. Liu & 2018 & DeepGauge: multi-granularity testing criteria for deep learning systems & ACM/IEEE International Conference on Automated Software Engineering \\
\cite{S52} & Zhang, M., Y. Zhang, L. Zhang, C. Liu and S. Khurshid & 2018 & DeepRoad: GAN-based metamorphic testing and input validation framework for autonomous driving systems. & ACM/IEEE International Conference on Automated Software Engineering\\
\cite{S53} & Guo, J., Y. Jiang, Y. Zhao, Q. Chen and J. Sun & 2018 & DLFuzz: differential fuzzing testing of deep learning systems & ACM Joint Meeting on European Software Engineering Conference and Symposium on the Foundations of Software Engineering\\
\cite{S54} & Rubaiyat, A. H. M., Q. Yongming and H. Alemzadeh & 2018 & Experimental Resilience Assessment of An Open-Source Driving Agent & arXiv preprint \\
\cite{S55} & Rhazali, K., B. Lussier, W. Schön and S. Geronimi & 2018 & Fault Tolerant Deep Neural Networks for Detection of Unrecognizable Situations & IFAC-PapersOnLine \\
\cite{S20} & Wicker, M., X. Huang and M. Kwiatkowska & 2018 & Feature-guided black-box safety testing of deep neural networks & International Conference on Tools and Algorithms for the Construction and Analysis of Systems (TACAS)\\
\cite{S65} & Linsley, D., D. Scheibler, S. Eberhardt and T. Serre & 2018 & Global-and-local attention networks for visual recognition & arXiv preprint \\
\cite{S3} & Wu, M., M. Wicker, W. Ruan, X. Huang and M. Kwiatkowska & 2018 & A Game-Based Approximate Verification of Deep Neural Networks with Provable Guarantees & arXiv preprint \\
\cite{S69} & Xu, K., D. H. Park, C. Yi and C. Sutton & 2018 & Interpreting Deep Classifier by Visual Distillation of Dark Knowledge & arXiv preprint \\
\cite{S56} & Mallozzi, P., P. Pelliccione and C. Menghi & 2018 & Keeping intelligence under control. & International Workshop on Software Engineering for Cognitive Services\\
\cite{S73} & Guidotti, R., A. Monreale, S. Ruggieri, D. Pedreschi, F. Turini and F. Giannotti & 2018 & Local rule-based explanations of black box decision systems & arXiv preprint\\
\cite{S57} & Guo, W., D. Mu, J. Xu, P. Su, G. Wang and X. Xing & 2018 & LEMNA: Explaining Deep Learning based Security Applications & ACM SIGSAC Conference on Computer and Communications Security\\
\cite{S71} & Tan, S., R. Caruana, G. Hooker, P. Koch and A. Gordo & 2018 & Learning Global Additive Explanations for Neural Nets Using Model Distillation & arXiv preprint \\
\cite{S72} & Dumitru, M. A. K.-R. M., E. B. K. S. D. Pieter, J. Kindermans and K. T. Schütt & 2018 & Learning how to explain neural networks: Patternnet and patternattribution & International Conference on Learning Representations (ICLR) \\
\cite{S31} & Xiang, W., H. D. Tran and T. T. Johnson & 2018 & Output Reachable Set Estimation and Verification for Multilayer Neural Networks & IEEE Transactions on Neural Networks and Learning Systems \\
\cite{S39} & Kuper, L., G. Katz, J. Gottschlich, K. Julian, C. Barrett and M. Kochenderfer & 2018 & Toward scalable verification for safety-critical deep networks & arXiv preprint \\
\cite{S46} & Cheng, C.-H., G. Nührenberg and H. Ruess & 2018 & Verification of binarized neural networks & arXiv preprint
\\
\hline
\end{longtable}
\end{footnotesize}

\end{document}